\documentclass[11pt]{article}

\usepackage[margin=3cm]{geometry}
\usepackage{caption}
\usepackage{subcaption}
\usepackage{graphicx}%
\usepackage{multirow}%
\usepackage{amsmath,amssymb,amsfonts}%
\usepackage{amsthm}%
\usepackage{mathrsfs}%
\usepackage{latexsym}%
\usepackage{mathtools}%
\usepackage[title]{appendix}%
\usepackage{xcolor}%
\usepackage{color,cancel}
\usepackage{textcomp}%
\usepackage{manyfoot}%
\usepackage[utf8]{inputenc}
\usepackage{listings}%

\usepackage{tocloft}
\usepackage{authblk}
\usepackage{lipsum}
\usepackage{wrapfig}
\usepackage{enumerate}
\usepackage{soul}
\usepackage{chemarrow}

\usepackage{booktabs}%
\usepackage{multirow}
\usepackage{rotating}
\usepackage[math]{cellspace}
\usepackage{algorithm}%
\usepackage{algorithmicx}%
\usepackage{algpseudocode}%
\usepackage{listings}%
\usepackage{units}
\usepackage{verbatim}
\usepackage{fancyvrb}
\fvset{fontsize=\normalsize}

\usepackage[T1]{fontenc}
\usepackage[sc]{mathpazo}
\setkeys{Gin}{width=\linewidth,totalheight=\textheight,keepaspectratio}

\usepackage{hyperref}
\hypersetup{
	hidelinks = true
}

\usepackage{cleveref}

\newcommand{\edit}[1]{#1}
\newcommand{\editmath}[1]{#1}

\usepackage{amsmath}
\usepackage{bm}
\makeatletter
\renewcommand*\env@matrix[1][*\c@MaxMatrixCols c]{%
  \hskip -\arraycolsep
  \let\@ifnextchar\new@ifnextchar
  \array{#1}}
\makeatother

\title{Enhancing generalizability of model discovery across parameter space with multi-experiment equation learning (ME-EQL)}
    
\author[1]{Maria-Veronica Ciocanel\thanks{\url{veronica.ciocanel@duke.edu}}}

\author[2]{John T. Nardini\thanks{\url{nardinij@tcnj.edu}}}

\author[3]{Kevin B. Flores}

\author[4]{Erica M. Rutter}

\author[4]{Suzanne S. Sindi}

\author[5]{Alexandria Volkening}

\affil[1]{Departments of Mathematics and Biology, Duke University, Science Drive, Durham, 27710, North Carolina, USA}

\affil[2]{Department of Mathematics and Statistics, The College of New Jersey, 2000 Pennington Road, Ewing, 08628, New Jersey, USA}

\affil[3]{Department of Mathematics, Center for Research in Scientific Computation, North Carolina State University, Raleigh, NC, USA}

\affil[4]{Department of Applied Mathematics, University of California, Merced, 5200 North Lake Road, Merced, 95340, California, USA}

\affil[5]{Department of Mathematics, Purdue University, 150 N.\ University St., West Lafayette, 47907, Indiana, USA}

\begin{document}

\maketitle
\begin{abstract}
Agent-based modeling (ABM) is a powerful tool for understanding self-organizing biological systems, but it is computationally intensive and often not analytically tractable. Equation learning (EQL) methods can derive continuum models from ABM data, but they typically require extensive simulations for each parameter set, raising concerns about generalizability. In this work, we extend EQL to Multi-experiment equation learning (ME-EQL) by introducing two methods: one-at-a-time ME-EQL (OAT ME-EQL), which learns individual models for each parameter set and connects them via interpolation, and embedded structure ME-EQL (ES ME-EQL), which builds a unified model library across parameters. We demonstrate these methods using a birth--death mean-field model and an on-lattice agent-based model of birth, death, and migration with spatial structure. Our results show that both methods significantly reduce the relative error in recovering parameters from agent-based simulations, with OAT ME-EQL offering better generalizability across parameter space. Our findings highlight the potential of equation learning from multiple experiments to enhance the generalizability and interpretability of learned models for complex biological systems. 
\end{abstract}

\begin{footnotesize}
\end{footnotesize}

\section*{Introduction}

Biological systems can exhibit rich spatiotemporal patterns and dynamics in which population-level behavior emerges from interactions at the level of individual entities, i.e., cells, organisms, molecules and animals \cite{wadkin2020recent,styles2021review,rangamani2007modelling,VolkeningRev}. Mathematical models can be used as quantitative tools for bridging our understanding of how interactions across multiple scales lead to emergent behaviors. Mechanistic mathematical models provide a powerful framework for investigating these biological systems, with model complexity and abstraction tailored to the biological question and computational constraints \cite{edelstein2005mathematical, murray2007mathematical, friedman2014mathematical}. Among mechanistic approaches, agent-based models (ABMs) are characterized by their high level of detail, i.e., with the capability to investigate biologically-relevant mechanisms and capture spatial effects by explicitly simulating individual agents and their interactions.  However, the computational demands of ABMs often limit their utility in large-scale inference tasks such as parameter estimation, uncertainty quantification, sensitivity analysis, and optimal design. To overcome these limitations, surrogate models consider aggregate or coarse-grained agent dynamics, enabling efficient computational exploration of parameter spaces while preserving key features that are causal to the underlying biological dynamics \cite{norton2025advances,fonseca2025optimal,nardini2024forecasting,jain2022smore,bergman2024connecting}.

The task of finding accurate surrogate models is challenging. Parameterized differential equation models are often preferable to ``black-box'' models, such as neural networks or Gaussian processes, because they offer interpretability, enable symbolic analytical techniques, and are compatible with established methods for parameter estimation and uncertainty quantification. Analytical methods exist to formally derive mean-field models from ABM simulations, such as moment closure methods \cite{kiss2017mathematics,plank2015spatial,johnston2012mean} and the Fokker-Planck equation \cite{Erban_Chapman_2020}. However, while mean-field approaches offer analytical and computational tractability, they generally rely on strong formal assumptions, such as the diminishing effect of higher-order correlations or homogeneous mixing. These assumptions may not hold for many ABMs of biological systems, which exhibit spatial structure, stochasticity, and heterogeneous interactions, making analytical derivations difficult to apply and validate \cite{baker_correcting_2010,nardini2021learning}. Moreover, the behavior of ABMs can vary significantly across parameter regimes, with spatial correlation effects emerging or becoming negligible depending on the specific parameter values. This variability motivates the need for frameworks to derive differential equation surrogate models that can robustly capture essential dynamics across a broad range of ABM parameter regimes \cite{gaskin_neural_2023, nardini2024forecasting}.
 
Equation learning (EQL) has emerged as a powerful tool for using time-series data to discover governing differential equations by identifying functions that accurately describe the rate of change of biological processes driving the system dynamics. EQL methods aim to solve the problem of symbolic regression, in which regression methods are used to find the mathematical expressions that best describe a model for fitting data. Symbolic regression encompasses genetic programming methods \cite{koza1994genetic,schmidt2009distilling} and neural network or statistical learning approaches \cite{brunton2016discovering,udrescu2020ai}, among others. For a recent review on the topic, see \cite{makke202interpretable}. One prominent approach in this field is the Sparse Identification of Nonlinear Dynamical Systems (SINDy) method, which relies on sparse regression techniques to identify the underlying equations governing complex systems \cite{brunton2016discovering}. While approaches like SINDy have predominantly been applied to uncovering governing equations of dynamical systems, their potential is expanding into other fields, including biology and ecology, where similar challenges in system identification exist \cite{mangan2016inferring, prokop2024from}. Recently, Nardini et al. proposed using EQL to automatically derive ODEs as surrogate models for ABMs of biological phenomena \cite{nardini2021learning}. While this approach successfully demonstrates the feasibility of learning interpretable mean-field models from ABM simulations, it was applied in a restricted setting in which surrogate models were only derived for fixed values of ABM parameters. In particular, each new parameter set requires retraining the surrogate model. This limitation restricts the ability of EQL to generalize across parameter spaces where emergent behaviors may change qualitatively. Recent advances in conditional equation learning and operator learning aim to address this by capturing parameter-to-dynamics mappings, but these methods often sacrifice interpretability or require latent representations that are difficult to interpret biologically \cite{lu2021learning}. These gaps highlight the need for new methods that can derive differential equations from ABMs in a way that generalizes across parameter regimes while preserving mechanistic insight.

\begin{figure}[t!]
\includegraphics[width=\textwidth]{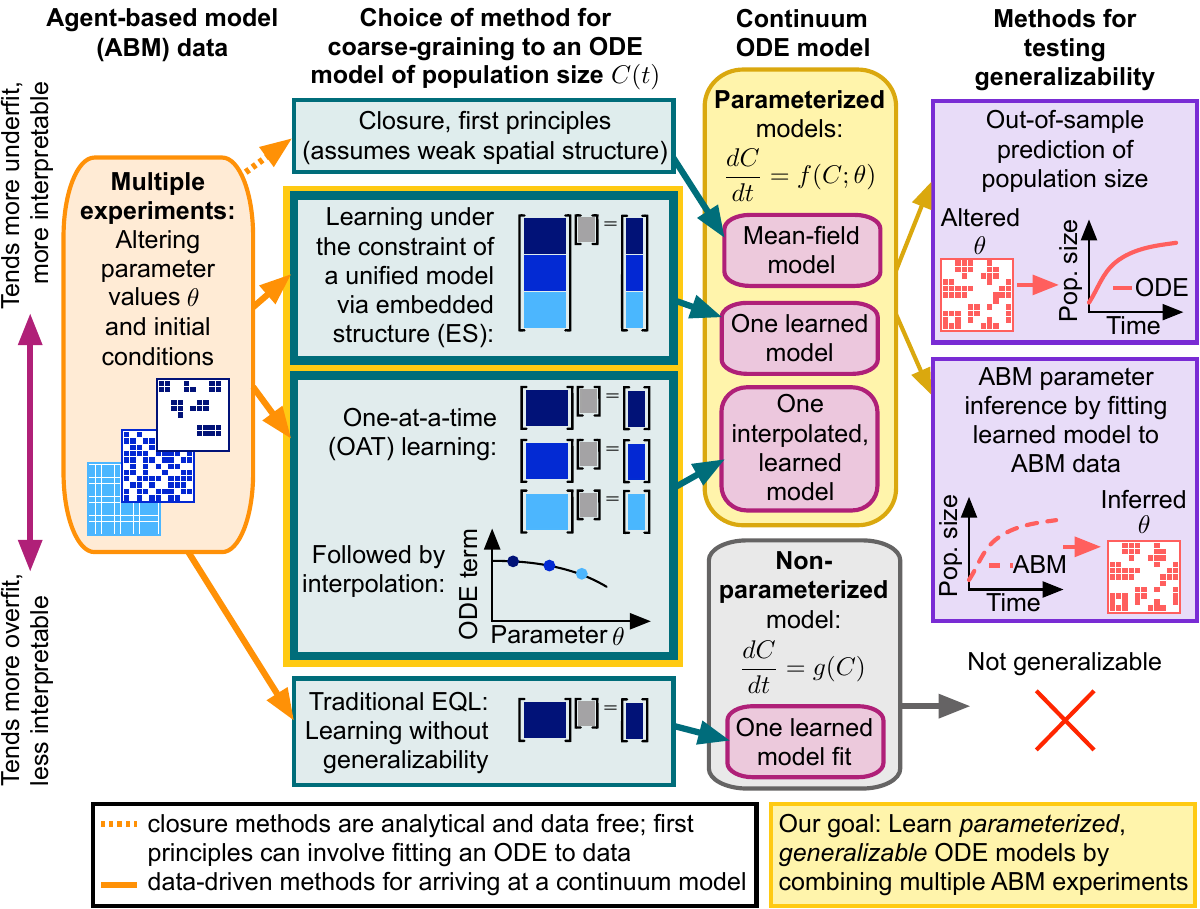}
\caption{\label{fig:conceptual}Overview of our motivation and approach. Agent-based models (ABMs) are a natural means of describing many biological systems, but these stochastic models often encounter challenges when researchers attempt analysis or parameter inference. Here, we describe new methods to utilize information from multiple experiments arising from different ABM parameter regimes (Orange). Traditional methods to develop coarse-grained models rely on closure assumptions that may lead to inaccurate representations of ABM spatial structure (Light Green, Top). Alternatively, traditional equation learning (EQL) methods involve discovering models from data, leading to excellent fits on training data but no means of generalizing to out-of-sample prediction (Light Green, Bottom). We propose two methods (Yellow) for addressing these challenges by performing EQL from multiple ABM experiments under different parameter values: ME-EQL. Our first method, ES ME-EQL, relies on learning ODEs from a library with embedded structure (ES) in the form of data and parameters from multiple ABM simulations; our second method, OAT ME-EQL, consists of repeating traditional EQL with different ABM parameters followed by interpolation to map these models to unobserved parameter values. Our approaches lead to parameterized ODEs (Pink), and we test their generalizability and interpretability by predicting ABM population size and inferring ABM parameter values (Purple).  }
\end{figure}

Here we develop two methods for learning generalizable differential equations from multiple ABM simulations conducted across different parameter values (i.e., a set of computational experiments). We refer to our methods collectively as multi-experiment equation learning or ``ME-EQL'' (Fig~\ref{fig:conceptual}). We consider ME-EQL approaches, including the two proposed here, as equation learning methods that also serve as a form of ``model discovery,'' because they recover explicit governing equations representing a unified mechanistic model shared across parameter regimes.
Our first approach, which we refer to as embedded structure multi-experiment EQL (ES ME-EQL), includes the biological parameters being varied explicitly as coefficients in the function library. Sparse regression is then applied to experiments from all observed parameters at the same time. In our second method, which we refer to as one-at-a-time multi-experiment EQL (OAT ME-EQL), we perform equation learning for each parameter choice separately and then interpolate coefficients over experiments to produce a model that generalizes across parameter space. 

Overall, we find that both ME-EQL methods exhibit significant promise for learning from a birth--death--migration model on a spatial lattice. This simple ABM is a canonical biological model, where the agents can represent cells in wound healing \cite{johnston_how_2014} or animal movement in ecology \cite{bernoff_nonlocal_2013}. From this ABM, one can derive equations that have been broadly used across biological systems, e.g., the Fisher-KPP equation \cite{fisher1937wave, swanson2003virtual, simpson2024fisher}. Moreover, this ABM and its extensions have previously been used to assess EQL performance \cite{nardini2021learning,nardini2024forecasting}. The success of these methods demonstrates the ability of learned differential equations to capture unmodeled effects such as spatial correlations and interactions between neighboring biological agents not explicitly included in the EQL library construction. We are not the first to consider the problem of learning across parameter space, as other studies have considered embedding the model parameters in a similar way \cite{nicolaou2023data}. However, to our knowledge, the work presented here is the first to compare different ME-EQL approaches and consider agent-based biological dynamics. This development of ME-EQL methodology is a prerequisite to applying EQL to more realistic biological scenarios in which no single differential equation form captures the full diversity of experimental conditions \cite{liu2024parameter}.


\section*{Methods and experiments}\label{sec:methods}

\subsection*{Generating data}\label{sec:datagen}

\textbf{Mean-field model.} Following \cite{nardini2021learning}, we used a classical birth--death (BD) model, a fundamental example in mathematical biology that describes a well-mixed population for an ABM on a lattice where individuals undergo birth, death and migration events. Agents proliferate (giving rise to a new agent) with rate  $R_p$ and die (and are removed from the lattice) with rate $R_d$. Invoking the mean-field assumption in this ABM produces a mean-field differential equation model for $C_{MFM}(t)$, which describes the evolution of the population density over time:
\begin{align}
    \frac{d}{dt} C_{MFM}(t) &= R_p C_{MFM}(t) \left( 1 - C_{MFM}(t) \right)  - R_d C_{MFM}(t) \\
    & = (R_p/2) C_{MFM}(t) - R_p C_{MFM}^2(t).
\label{eqn:MFM}
\end{align}
To simplify our analysis, we set $R_d = R_p/2$ and vary $R_p = \{0.01, 0.02, \dots ,5\}$. We consider two initial conditions: $C_{MFM}(0) = 0.05$ and $C_{MFM}(0) = 0.25$, which we denote IC=0.05 and IC=0.25, respectively. 

We simulate data $C_d(t)$ under differing amounts of noise:
\begin{equation}
    C_d(t) = C_{MFM}(t) +  \overline{C_{MFM}(t)} \mathcal{E} 
\end{equation}
\noindent
where $\overline{C_{MFM}(t)}$ represents the mean value of $C_{MFM}(t)$ and $\mathcal{E} \sim N(0,\sigma)$ represents i.i.d. Gaussian noise with standard deviation $\sigma$. We examine two cases of the MFM under differing noise levels: (1) No noise ($\sigma =0$) and (2) low noise ($ \sigma = 0.0025$).

\vspace{2mm}
\noindent\textbf{Agent-based model.} We generated synthetic data from an ABM for the scenario where no differential equation model is known to accurately approximate the ABM for all parameters. In the ABM, each agent exists on a two-dimensional lattice with reflecting boundary conditions. We assume that each agent can proliferate with rate $R_p$, dies with rate $R_d = \frac{R_p}{2}$, and migrates to an adjacent lattice site at the rate $R_m  = 1$. Unlike the MFM model, the ABM captures spatially-heterogeneous interactions, but this spatial structure is \textit{hidden} in our observations because we only track the population density over time. Moreover, this ABM has been previously shown to be approximated using the same mean-field model as Eq~\eqref{eqn:MFM} in some parameter regimes \cite{baker_correcting_2010,nardini2021learning}. This allows us to compare the learned equations in different parameter regimes, where the mean-field assumption may or may not be accurate. For further details about simulating the ABM, see \cite{nardini2021learning}.

Each simulation is independently initialized by selecting 5\% or 25\% of the lattice sites uniformly at random  for occupation. We denote these initial conditions as IC = 0.05 and IC = 0.25, respectively. For each choice of proliferation rate $R_p$, we generate 25 independent simulations, and for each simulation we track the total density of occupied sites over time:
$$C^{(i)}_{ABM}(t) = \frac{T^{(i)}(t)}{X^2}\,,$$
where $X^2 = 120^2$ is the size of the lattice and $T^{(i)}(t)$ is the total number of occupied sites in simulation $i$ and at time $t$. The data we consider for equation learning, $C_d(t)$, is the average of all ABM simulations:
\begin{equation}\label{eq:mean_ABM}
    C_d(t) = \left< C_{ABM}(t)\right> = \frac{1}{N} \sum_{i = 1}^{N} C^{(i)}_{ABM}(t),
\end{equation}
\noindent
where $N=25$ is the number of simulations averaged for each $R_p$ value.

\subsection*{Equation learning}\label{sec:EQL}

The goal of an equation learning (EQL) framework is to learn the dynamical systems model given by 
\begin{equation}
    \dfrac{dC(t)}{dt} = \mathcal{F} \label{eq:EQL_ODE_RHS}
\end{equation}
that best describes observations of the dynamics, $C_d(t)$. Note that for simplicity of notation, we do not include subscripts to denote the time points at which data are observed, i.e., $C_d(t)$ is short-hand notation for data collected at times $\{t_i\}_{i=1}^n$ corresponding to $\{C_d(t_i)\}_{i=1}^n $, which might contain observation or process noise. Our EQL method builds on the SINDy (Sparse Identification of Nonlinear Dynamics) methodology \cite{brunton2016discovering}. The general approach is that a library of potential terms is created for $\mathcal{F}$, and sparse regression is used to select the most parsimonious model that describes the data. In the following, we discuss the steps involved in EQL: Step (1) approximating the time derivative of the data, Step (2) constructing the library, and Step (3) sparse regression for model selection.

\vspace{2mm}

\noindent
\textbf{Step 1: Derivative approximation from data.} To find the appropriate right-hand side of Eq~\eqref{eq:EQL_ODE_RHS}, we must calculate $\frac{dC_d(t)}{dt}$ using data $C_d(t)$. 
Previous studies have shown that the presence of noise in the observed data can be amplified when using finite-differencing to calculate derivatives \cite{lagergren2020learning}. To account for this, we use \texttt{smoothdata} in Matlab to smooth the derivatives obtained using forward finite differences for our ABM data. In the case of the MFM data, we control the (small) amount of noise added to the data and numerically approximate the derivatives using the \texttt{numpy.gradient} function for central finite differences in Python.

\vspace{2mm}

\noindent
\textbf{Step 2: Library construction.} The library of potential right-hand side terms is constructed by forming a matrix $\Theta$ in which the rows correspond to time points and columns correspond to library terms evaluated at those time points.
In this manuscript, we use polynomial terms. However, any functional forms that the user postulates are important to explain the underlying dynamical system that generated the data could be included (e.g., trigonometric or exponential functions).

\vspace{2mm}

\noindent
\textbf{Step 3: Sparse regression.} For a library of model terms $\Theta$, and time derivatives of the data $\frac{dC_d(t)}{dt}$, sparse regression is applied to the linear equation defined by \begin{equation}
    \dfrac{dC(t)}{dt} = \Theta\xi \label{eq:SINDy_eqn}
\end{equation} to estimate a sparse vector of parameters, $\xi$, found by solving the optimization problem \begin{equation}
\hat{\xi} = \mathop{\arg \min}\limits_{\xi} \left|\left|\frac{dC_d}{dt} - \Theta \xi \right|\right|_2^2 + \lambda\left| \xi \right|_1.
\label{eq:minimizer}
\end{equation}
The $\ell_1$ penalty added to the objective function promotes sparsity and the hyperparameter $\lambda$ is used to tune overfitting \cite{tibshirani_regression_1996}. We use cross-validation and Akaike Information Criteria (AIC) scores to select the optimal $\lambda$, as is used in existing literature \cite{mangan2017model}. While the details can be found in the hyperparameter selection section in \nameref{S1_file}, we briefly describe the process below:
\begin{itemize}
\item Randomly split the data into 10 training and testing sets, each with 80\% training and 20\% validation, denoted $C_{d,k},$ where $k=1,..,10$ denotes the train-test splits.
\item For each train-test split, we solve \Cref{eq:minimizer} using the pysindy package in python \cite{desilva2020,Kaptanoglu2022} over a grid of 100 equi-log-spaced values $\lambda_j = 10^{-1},...,10^{-9}$ to obtain the optimal coefficients $\xi_{k,j}$.
\item For each $\xi_{k,j}$, we forward solve \Cref{eq:SINDy_eqn} and calculate the AIC score. At each $\lambda_j$ value, we average the 10 AIC scores from the test-train splits to obtain $\bar{\lambda}_j$.
\item The final $\lambda$ value is selected by being the smallest $\bar{\lambda}_j$ while also satisfying that $|\xi_{j,k}|$ \edit{is below a pre-defined threshold} for all 10 test-train splits. The first criteria ensures a parsimonious fit while the second criteria leverages domain knowledge that \edit{coefficients} should not be overly large (see \nameref{S1_Figure} to \edit{visualize} AIC \edit{score} versus $\bar{\lambda}$).
\end{itemize}

\vspace{2mm}

\noindent \textbf{Multi-experiment equation learning (ME-EQL).}
We consider two learning approaches for enhancing generalizability to parameter sets not used in the training data for equation learning. We perform the optimization problem defined by Eq~\eqref{eq:minimizer} over data arising from multiple experiments. We refer to the first approach as ``one-at-a-time'' (OAT ME-EQL) (parameter-specific/individual experiment) and to the second approach as ``embedded-structure'' (ES ME-EQL). Step 1 (approximating the time derivative from data) is the same for each method, although the approaches diverge in Steps 2 and 3. 

\vspace{2mm}

\noindent
\textbf{OAT ME-EQL.} In this approach, we learn the underlying dynamics independently for each dataset generated using parameter $R_p$. In Step 2, the library of potential terms is $\Theta = [C, C^2, \dots, C^{10}].$ In Step 3, the hyperparameter $\lambda$ is selected for each $R_p$ dataset independently (see hyperparameter selection section in the \nameref{S1_file} for more details). \edit{In step 4, the threshold for $|\xi_{j,k}|$ is set to 100}. This results in separate $\hat{\xi}$ learned for each $R_p$ in Eq~\eqref{eq:minimizer}. This single-experiment learning (from each individual $R_p$) uses SINDy and hyperparameter selection, and we refer to it as \textbf{OAT EQL} (one-at-a-time EQL). We thus learn potentially different model structures and coefficients for the datasets corresponding to each parameter value. We note that our approach of combining information from multiple simulations shares similarities with previous ensemble-style equation learning methods, i.e., Ensemble-SINDy \cite{fasel2022ensemblesindy}. However, our OAT-ME-EQL method differs because it utilizes data generated from distinct parameter values, whereas Ensemble-SINDy uses simulations generated by the same fixed parameter values but with varied initial conditions. Interpolation across coefficients of the most commonly-learned right-hand-side terms is performed, yielding a single equation that generalizes across all parameters $R_p$. See Algorithm~1 in \nameref{S2_file} and \nameref{sec:genmethods} for more details. 

\vspace{2mm}

\noindent
\textbf{ES ME-EQL}. In this approach, a single model is learned for all parameter values $R_p$ and over all datasets jointly. To accomplish this, the library of potential terms in Step 2 is assumed to be $\Theta = [R_p C, R_p C^2, \dots, R_p C^{10}].$ In other words, we assume that the coefficients in the learned ODE are linear in $R_p$; this simple parameter dependence is inspired by the mean-field model~\eqref{eqn:MFM}. Similarly, the hyperparameter $\lambda$ is selected jointly over all datasets to ensure that only one model is learned. \edit{In step 4, the threshold for $|\xi_{j,k}|$ is set to 20}. Thus, in this case, a single $\hat{\xi}$ is learned corresponding to only one model structure that is learned for all $R_p$ values. See Algorithm~2 in \nameref{S2_file} and \nameref{sec:genmethods} for more details.

\subsection*{Generalizability over parameter space}
\label{sec:genmethods}
A key aim of our framework is to enable generalization to parts of parameter space that were not used to learn the differential equation (DE) models. In each of the equation learning methods we consider, hyperparameter tuning and model selection occurs using model simulations from a select number of parameter sets (e.g., 5 or 10 separate $R_p$ values, which we refer to as experiments). Below, we describe our methodology for using learned DE models to estimate parameters from model simulations that were not included in the training data set.

\vspace{2mm}

\noindent
\textbf{OAT ME-EQL.} In this framework, it is possible that different models structures are learned for each dataset corresponding to a different value of the parameter $R_p$. Model structure is defined as the collection of library terms with non-zero coefficients. To enable generalization to unseen parameters, we select the most common model structure given by:
\begin{equation}
    \frac{dC}{dt}=\xi_1C+\xi_2C^2+\dots+\xi_{10}C^{10},
    \label{eq:veronicamodel}
\end{equation}
which holds for the majority of $R_p$ values that resulted in the same set of non-zero coefficients $\xi_i$. The coefficient vectors $\boldsymbol{\xi_i}, i =1,\dots,10$, contain the parameters $\xi_i$ learned for all $R_p$ values. We then interpolate coefficients across parameter space using cubic splines or lines to obtain a function $\xi_i(R_p)$. We only use the most common model for interpolation. For example, if we select 10 $R_p$ values, but only 8 have the same model structure, we would only use those 8 coefficient sets for interpolation. The generalized model is then given by:
  \begin{equation}
    \frac{dC}{dt}=\xi_1(R_p)C+\xi_2(R_p)C^2+\dots+\xi_{10}(R_p)C^{10}.
    \label{eq:interpyderp}
\end{equation}
Note that if there is no dominant common model structure, this provides insight into whether it is possible for a single differential equation model from our library to generalize across parameter space.

\vspace{2mm}

\noindent
\textbf{ES ME-EQL.} Since one unified model is learned over all parameter sets and $R_p$ is incorporated into the library, generalization to outside parameter sets is trivial. Our learned model has the structure:
\begin{equation}
    \frac{dC}{dt}=\xi_1R_pC+\xi_2R_pC^2+\dots+\xi_{10}R_pC^{10}.
    \label{eq:johnmodel}
\end{equation}
Thus, to predict $C(t)$, Eq~\eqref{eq:johnmodel} is simulated at a given out-of-sample parameter value of $R_p$.\\

\section*{Results}
We test both ME-EQL frameworks (OAT ME-EQL and ES ME-EQL) with increasingly complex data. First, we examine the ability of the proposed algorithms to generalize across parameters when the data is generated using a known underlying model in \nameref{sec:results_meanfield}. We examine how the methods perform when increasing noise or decreasing information content, which we control by altering the initial condition to reduce the dynamic range of population densities observed (see \nameref{S2_Figure}). Secondly, we investigate algorithm performance in a situation where there may not be a known underlying model by using ABM data output in \nameref{sec:results_agentbased}. Lastly, in \nameref{sec:inferRP}, we examine the ability to recover ABM parameters from a single ABM simulation using our learned equations.

We also compare the results from both ME-EQL frameworks for these datasets with two established methods: (1) mean-field model approximations and (2) EQL for one parameter at a time, with no interpolation. Recall that we refer to this second method as OAT EQL (see \nameref{sec:EQL}), since it finds an equation for each $R_p$ value but is not intended to generalize its results between $R_p$ values.

\subsection*{Learning for noisy mean-field model data} \label{sec:results_meanfield}

\begin{figure}[t!]
\centering
\includegraphics[width=0.5
\textwidth]{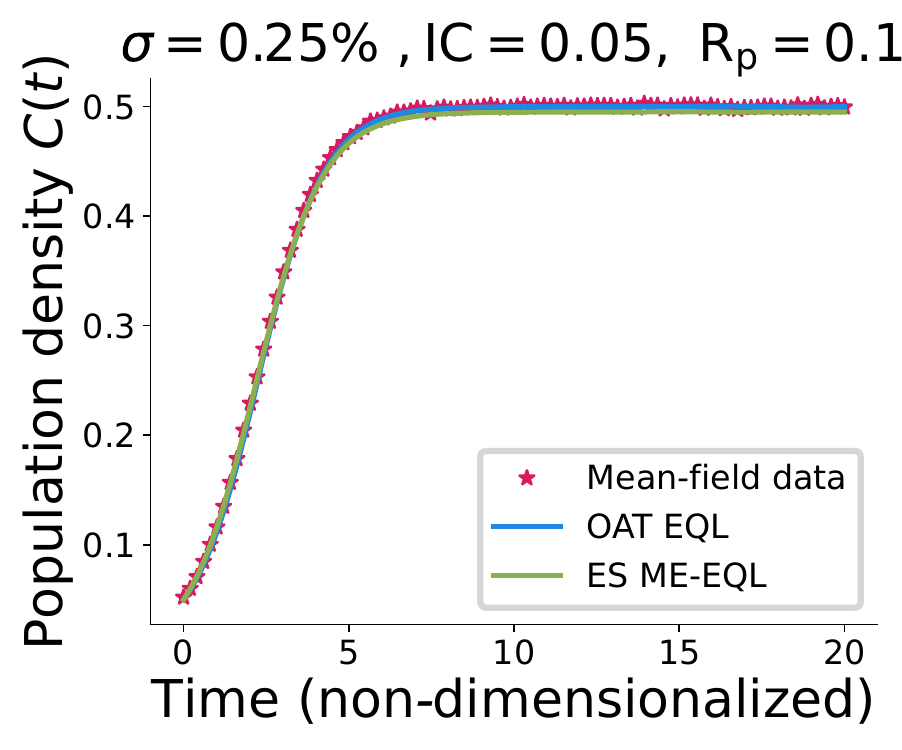}
\caption{Sample dataset generated using the Mean-Field Model (Eq~\eqref{eqn:MFM}) with added 0.25\% proportional noise (black stars) and fits with the OAT EQL approach (blue line) and the ES ME-EQL approach (green line). The sample MFM dataset shown here is generated using proliferation rate $R_p=0.1$ and initial condition IC=0.05. \label{fig:example_fitl}}
\end{figure}

We first consider data generated using the mean-field model Eq~\eqref{eqn:MFM} under no noise or low noise ($\sigma = 0.25\%$). We denote this by MFM data. To assess the impact of information content on the learned model and generalizations, we also examine two initial conditions: IC=0.05 and IC=0.25. As we show in \nameref{S2_Figure}, shifting from IC=0.05 to IC=0.25 effectively means that we observe a much narrower range of the population dynamics, leading to less information content. (For example, notice that if we were to consider the extreme case of IC=0.5, the mean-field model would be at equilibrium and there would be no information---other than the equilibrium value---available in the data for equation learning.) Fig~\ref{fig:example_fitl} displays the comparison between a single MFM dataset with $0.25\%$ noise (black stars) and the resulting OAT EQL fit (blue line) and ES ME-EQL fit (green line) for proliferation rate $R_p=0.1$. These results indicate that the learned models accurately fit the simulated data. 

\begin{figure}[t!]
\centering
\includegraphics[width=
\textwidth]{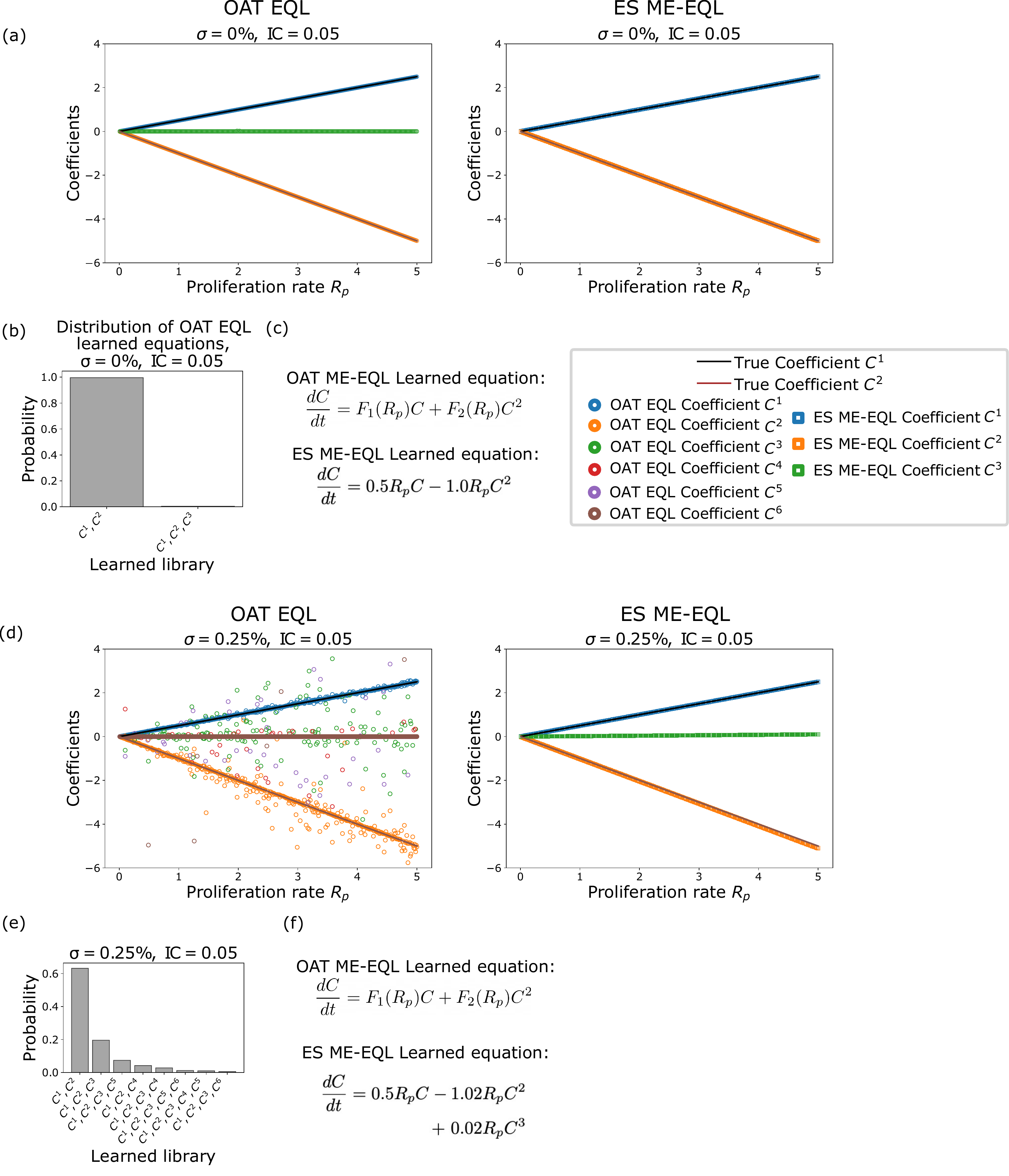}
\caption{Learned coefficients and models for the mean-field model Eq~\eqref{eqn:MFM} for IC = 0.05 with no noise (top) and 0.25\% noise (bottom). Panels (a), (d) display the true model coefficients (black lines), learned model coefficients using OAT EQL (colored circles), and learned model coefficients using ES ME-EQL (hollow shapes). Panels (b), (e) depict histograms of the  frequencies of the learned models for OAT EQL. Panels (c), (f) list the learned OAT ME-EQL and ES ME-EQL models. In the noise-free} case (a-c), both OAT ME-EQL and ES ME-EQL learn the model coefficients accurately. The recovered model for ES ME-EQL and the most commonly recovered model for OAT EQL ($99.6\%$) is the true underlying model in Eq~\eqref{eqn:MFM}. For the case with noise (d-f), the recovered coefficients do not always match the known model coefficients. However, $63\%$ of the learned OAT EQL models recover the true underlying model Eq~\eqref{eqn:MFM}. ES ME-EQL recovers a small extra cubic term. 
\label{fig:mean-field_EQL}
\end{figure}

First, we examine the models learned using our ME-EQL frameworks using data simulated at proliferation rate values $R_p=\{0.01,0.02,\dots, 5\}$. Fig~\ref{fig:mean-field_EQL}a shows that there is clear agreement between the true coefficients (black lines), the learned OAT EQL coefficients (circles), and the learned ES ME-EQL coefficients (squares and triangles). Over the 500 $R_p$ values used to generate MFM data, the OAT EQL method learns the correct underlying model in 498 cases, while the other two cases learn a structure that incorporates a $C^3$ term (Fig~\ref{fig:mean-field_EQL}b). The recovered underlying model for the ES ME-EQL and OAT ME-EQL approaches is the correct model (Fig~\ref{fig:mean-field_EQL}c). 

When adding small levels of noise ($\sigma = 0.025\%$) to the generated data, there are clear impacts for the recovered models for both the OAT EQL and ES ME-EQL approaches. Fig~\ref{fig:mean-field_EQL}d indicates that the learned coefficients do not always reflect the correct underlying model, as in the noise-free case. For example, in the OAT EQL recovered coefficients, there are not only $C^1$ (blue circles) and $C^2$ (orange circles) coefficients, but sometimes coefficients for $C^3,\dots,C^6$ terms. While many of the recovered coefficients agree with the true coefficients (black line), there are some deviations. For the ES ME-EQL recovered coefficients, we learn coefficients for $C^1$ (squares), $C^2$ (triangles), and $C^3$ (diamonds). The $C^1$ coefficients match well with the underlying true coefficient values, however there are slight deviations in the $C^2$ coefficients, especially as $R_p$ increases. The learned $C^3$ coefficients are small, but no such term exists in the underlying model. OAT EQL learns a larger variety of model structures (Fig~\ref{fig:mean-field_EQL}e) with this small amount of added noise, but identifies the correct model for over 60\% of the $R_p$ values. ES ME-EQL thus does not learn the correct model (given the small $C^3$ term), while the OAT ME-EQL method most commonly learns the correct model (Fig~\ref{fig:mean-field_EQL}f).

To assess the generalizability of the methods, we apply the learning frameworks to a smaller set of the data, generated using only 10 or 5 $R_p$ values. When learning from 10 experiments, we select $R_p = [0.01, 0.51, 1.01, \dots, 4.51]$ and when learning from 5 experiments, we select $R_p = [0.01, 1.01,\dots,4.01]$. As described in \nameref{sec:genmethods}, we generalize the learned equations to unseen $R_p$ values. The learned models can be found in \nameref{S1_Table}. We then compare the mean squared error (MSE) between the generalized recovered model and the noisy ABM data (Eq~\eqref{eq:mean_ABM}) corresponding to each $R_p$ parameter.

\begin{figure}[t!]
\centering
\includegraphics[width=
\textwidth]{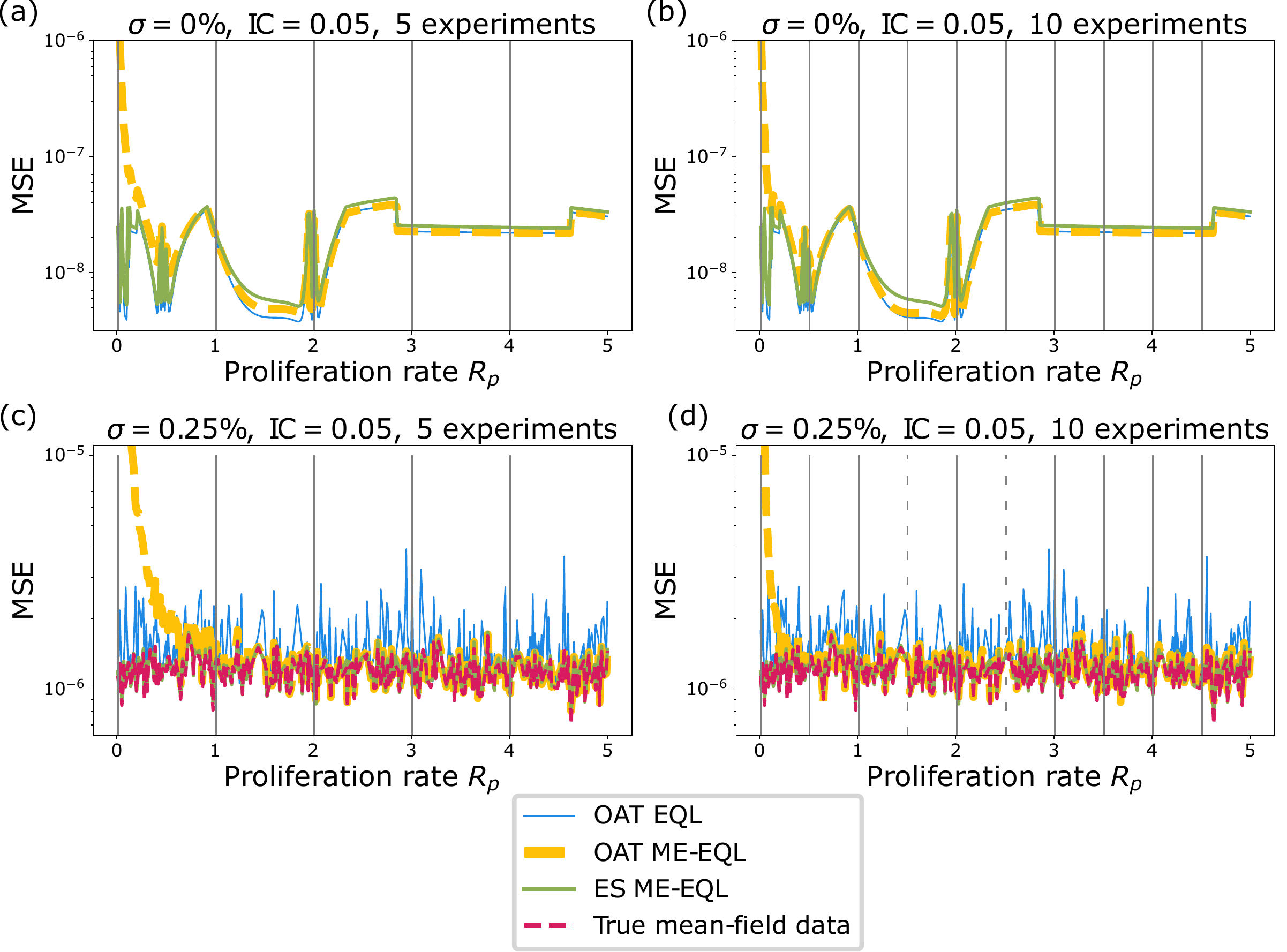}
\caption{MSE between data and recovered models for 0\% noise (a-b) and 0.25\% noise (c-d). The results from OAT EQL from each separate $R_p$ value are shown in blue for comparison purposes, the OAT ME-EQL learning is shown in yellow dashes, and the ES ME-EQL learning is shown in green solid lines. The gray vertical bars indicate the small set of $R_p$ values from which coefficients were learned from. Dashes indicate that OAT ME-EQL did not include the dataset corresponding to that $R_p$ value, since this framework did not learn the most popular model at that parameter value. In panels (a), (c), OAT ME-EQL and ES ME-EQL learn from maximum 5 $R_p$ values, and in panels (b), (d), OAT ME-EQL and ES ME-EQL learn from maximum 10 $R_p$ values. The red dashed lines represent the error added to the MFM model, which is only shown in the noisy case. }
\label{fig:mean-field_EQL_interp}
\end{figure}

In the case of noise-free data, it is clear that all methods perform similarly in terms of MSE, except when $R_p$ values are small (Fig~\ref{fig:mean-field_EQL_interp}a,b). Surprisingly, even with as few as 5 experiments (Fig~\ref{fig:mean-field_EQL_interp}a), the correct underlying model is learned for both the OAT ME-EQL and ES ME-EQL methods. The interpolation introduces little error in the OAT ME-EQL method, except when proliferation rates are small. As we increase from 5 to 10 experiments for interpolation, the region of $R_p$ space with higher MSEs slightly decreases (Fig~\ref{fig:mean-field_EQL_interp}b). In the case of noisy data, both OAT ME-EQL and ES ME-EQL appear to have MSE values that are on par with the known underlying noise level (Fig~\ref{fig:mean-field_EQL_interp}c,d). However, the ES ME-EQL method learns an additional $C^3$ term. Despite this, the recovered MSE values are small. We also find that OAT ME-EQL learns the correct model even with the introduction of noise. In general, all methods recover MSEs on the same order of magnitude, resulting in similar forward solutions (see Fig~\ref{fig:example_fitl}).

\begin{figure}[t!]
    \centering
    \includegraphics[width=0.95\linewidth]{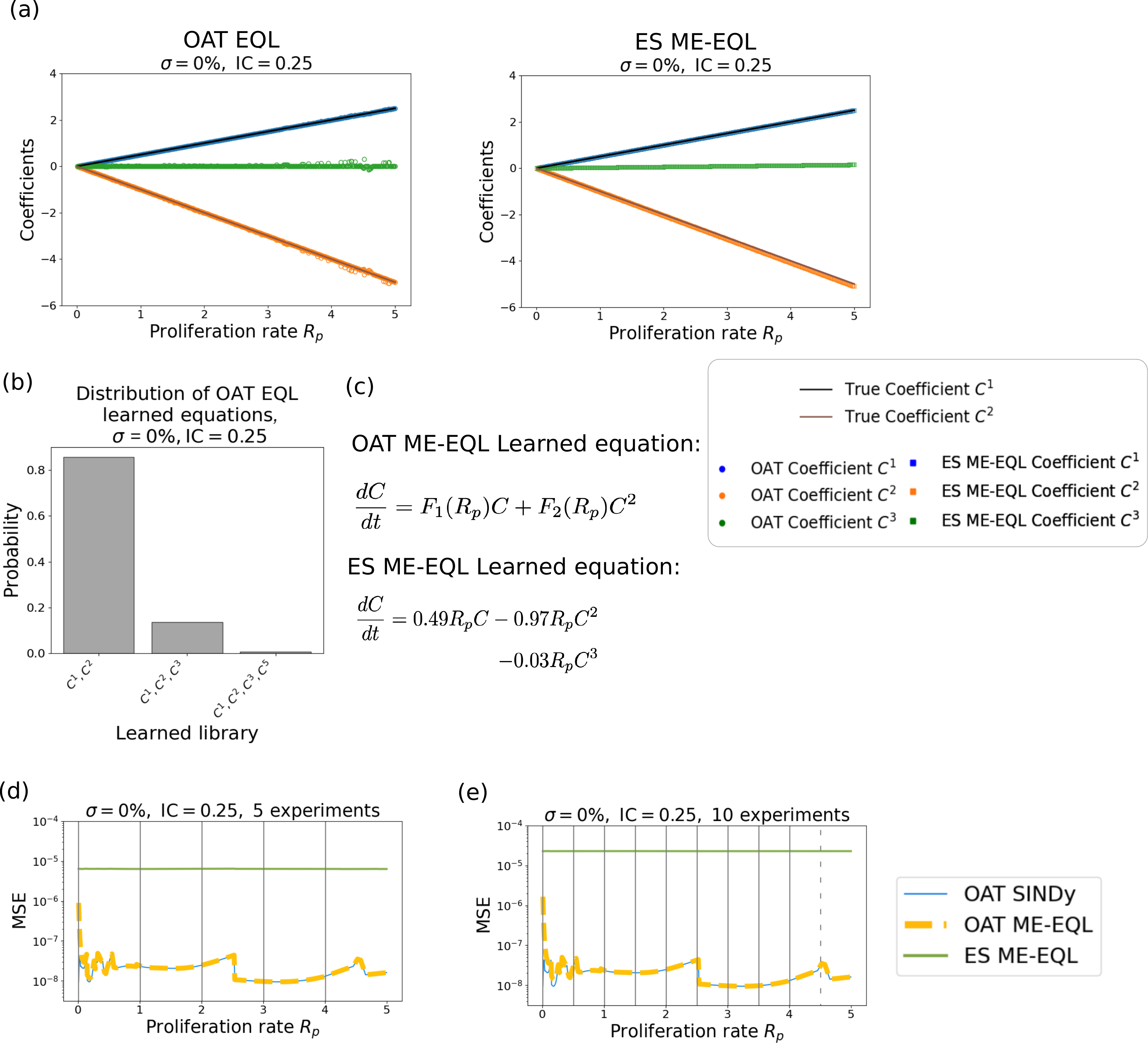}
    \caption{ Mean-field learning with IC = 0.25 and $\sigma=0\%$. (a) Learned Equations using the ES ME-EQL and OAT EQL approaches. (b) Most common learned equations from the OAT EQL approach. (c) Learned equation from the ES ME-EQL approach. (d) MSE of ME-EQL frameworks in predicting mean-field data over all $R_p$ values using 5 experiments. (e) MSE of ME-EQL frameworks in predicting mean-field data over all $R_p$ values using 10 experiments. }
    \label{fig:mean-field_EQL_025}
\end{figure}

We also examined the learning outcomes when using the larger initial condition of IC=0.25 (Fig~\ref{fig:mean-field_EQL_025}). The ES ME-EQL learned model contains an extra $C^3$ term, even when the data is noise-free  (Fig~\ref{fig:mean-field_EQL_025}c). However, the learned coefficients in the ES ME-EQL model are similar to the known underlying coefficients and the coefficient in front of the $C^3$ term is small. The learned models using OAT EQL are mostly the correct model form ($>80\%$, Fig~\ref{fig:mean-field_EQL_025}b), although higher order terms are learned for a small proportion of $R_p$ values.

In terms of generalizability, the OAT ME-EQL approach outperforms ES ME-EQL. When using 5 experiments for interpolation (Fig~\ref{fig:mean-field_EQL_025}d), the MSEs are several orders of magnitude smaller than those from ES ME-EQL. This is likely due to ES ME-EQL learning the incorrect model with slightly different coefficients (see \nameref{S1_Table}). We observe similar results when learning from 10 experiments (Fig~\ref{fig:mean-field_EQL_025}e). Compared to the IC=0.05 case, we find that (1) equation learning more frequently fails to capture the underlying model in the OAT EQL case, and (2) ES ME-EQL does not always recover the true model even when the data is noise-free. We suspect that this is due to the smaller information content in the data corresponding to the IC=0.25 initial condition. 

Overall, when the underlying model is known, we find that both methods perform similarly well in recovering models that match the data with small residuals. The ES ME-EQL approach has the advantage that it is required to learn a single equation, which is often the true model, and performs well on out-of-sample experiments. There are, however, settings where it learns an additional term of the ODE model, especially when confronted with more noise and less information content. In contrast, the OAT ME-EQL approach uses the most popular learned equation and interpolates the coefficients across parameter values. In general, the most popular recovered model matches the known underlying dynamical system, however the interpolated coefficients do not extrapolate well to small $R_p$ values. Both methods perform well when learning from as few as 5 experiments.

\subsection*{Learning for agent-based model data} \label{sec:results_agentbased}

\begin{figure}[t!]
\centering
\includegraphics[width=
\textwidth]{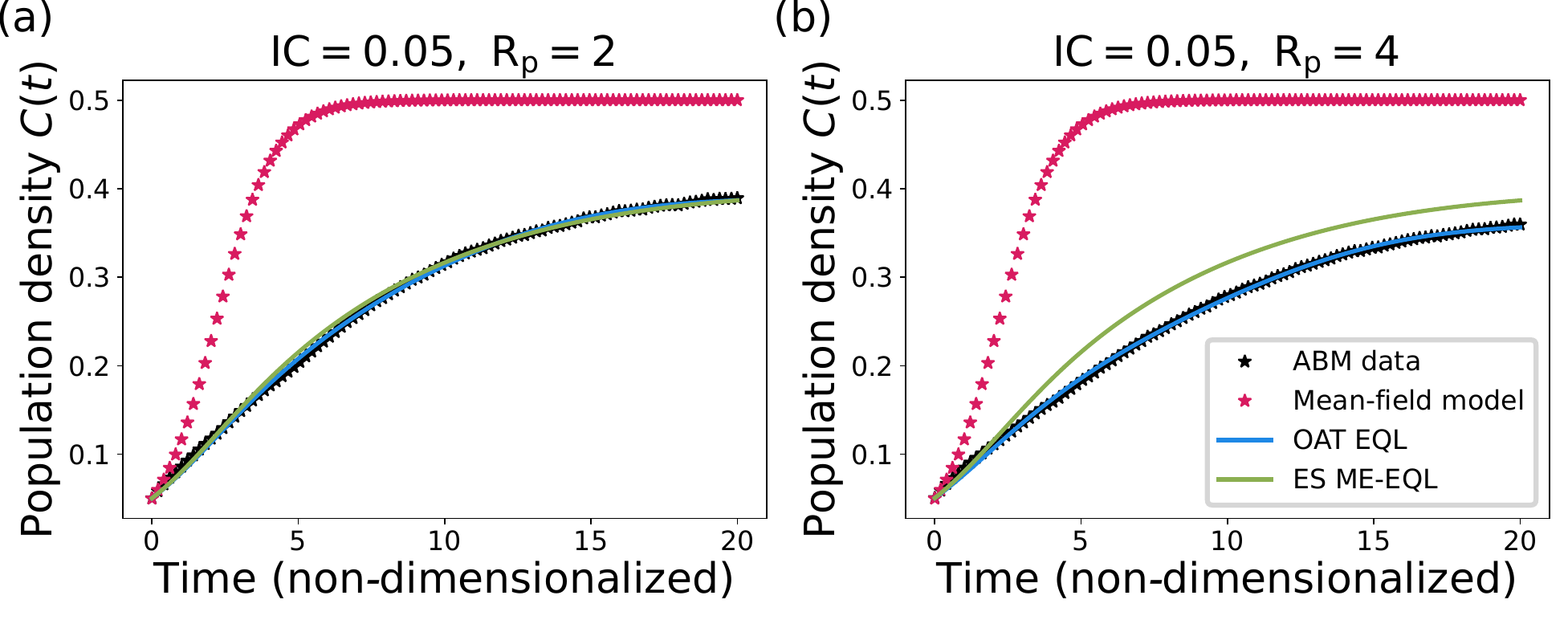}
\caption{Sample datasets generated using the ABM model (black stars) and fits with the OAT EQL approach (blue line) and the ES ME-EQL approach (green line). The ABM datasets shown here are generated using proliferation rate $R_p=2$ (a) and  $R_p=4$ (b) and initial condition IC=0.05.\label{fig:ABM_example_fits}}
\end{figure}

To assess generalizability in cases where there is no known underlying dynamical system, we apply OAT ME-EQL and ES ME-EQL to the ABM data generated as described in \nameref{sec:datagen}. Fig~\ref{fig:ABM_example_fits} displays sample ABM data (black stars) for two $R_p$ values, as well as the corresponding mean-field DE models (red stars), and the recovered OAT ME-EQL (blue) and ES ME-EQL (green) models. For both $R_p$ values, the mean-field approximation represents a poor match to the ABM data, demonstrating that we no longer have a ``ground-truth" model when learning equations for this ABM data, due to the spatial heterogeneities in the model. We aim to assess whether the two ME-EQL methods can learn ODE models that describe the ABM behavior across parameter space, and to identify parameter regimes where the mean-field model is insufficient to describe the dynamics.

\begin{figure}[t!]
\centering
\includegraphics[width=
\textwidth]{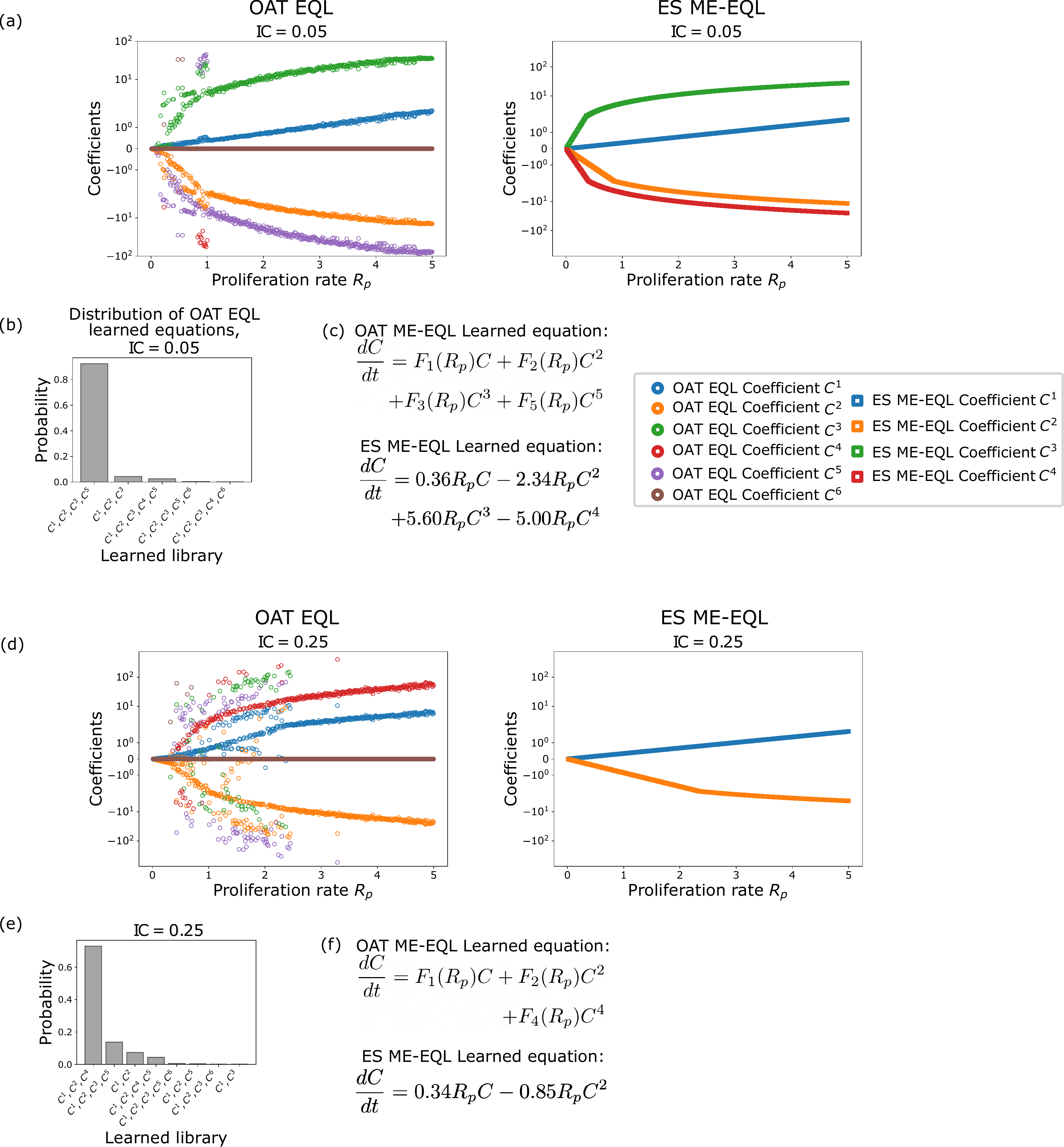}
\caption{Learned coefficients and models for the agent-based model for IC = 0.05 (top) and IC = 0.25  (bottom).  Panels (a), (d) display the learned model coefficients using OAT EQL (colored circles) and learned model coefficients using ES ME-EQL (hollow shapes). Panels (b), (e) depict a histogram of the frequencies of the learned models for OAT EQL. Panels (c), (f) list the learned OAT ME-EQL and ES ME-EQL models. For the IC=0.05 case (a-c), there is greater agreement between the OAT EQL and ES ME-EQL learned coefficients, although there are deviations for OAT EQL when $R_p<0.5$. The most commonly-learned model for OAT EQL ($92\%$) contains four terms (up to $C^5$), while the ES ME-EQL learned model contains four terms (up to $C^4$). For the IC=0.25 case (d-f), there is greater disagreement between the learned coefficients from OAT EQL and ES ME-EQL. There are more deviations from continuity for OAT EQL coefficients and over a larger set of $R_p$ space. The recovered model for ES ME-EQL is logistic, while the most commonly-recovered model for OAT EQL ($73\%$) contains three terms (up to $C^4$).} \label{fig:ABM_coeffs_probs}
\end{figure}

Fig~\ref{fig:ABM_coeffs_probs} displays the learned coefficients, distribution of learned equations, and final learned equations for the OAT EQL, ES ME-EQL and OAT ME-EQL methods for IC = 0.05 (top) and IC = 0.25 (bottom). For IC = 0.05, the learned coefficients show some continuity with small variation for  OAT EQL, and there are clearly differences between the coefficients learned with OAT EQL and ES ME-EQL (Fig~\ref{fig:ABM_coeffs_probs}a). The most commonly-learned OAT EQL equation contains four terms (up to $C^5$) while the ES ME-EQL equation learns four terms (up to $C^4$) (Fig~\ref{fig:ABM_coeffs_probs}b,c). There is more noise in the learned coefficients for IC = 0.25, which is due in part to the additional model structures learned with OAT EQL (Fig~\ref{fig:ABM_coeffs_probs}e). The most commonly-learned model structure for OAT EQL is learned for 73\% of $R_p$ values (as compared to 92\% for IC = 0.05). This model structure contains 3 terms (up to $C^4$) for OAT ME-EQL, while the ES ME-EQL learned equation recovers a logistic function (Fig~\ref{fig:ABM_coeffs_probs}f). 

The ES ME-EQL methodology learns fourth-order and second-order models for initial conditions 0.05 and 0.25, respectively. Interestingly, the model structure learned by ES ME-EQL for the initial condition of 0.05 is not learned for any single parameter set in OAT EQL. Moreover, for  IC = 0.25, ES ME-EQL learns the model structure of the mean-field model, but with different parameter values. The parameterized learned models can be found in \nameref{S2_Table}.

\begin{figure}[t!]
\centering
\includegraphics[width=
\textwidth]{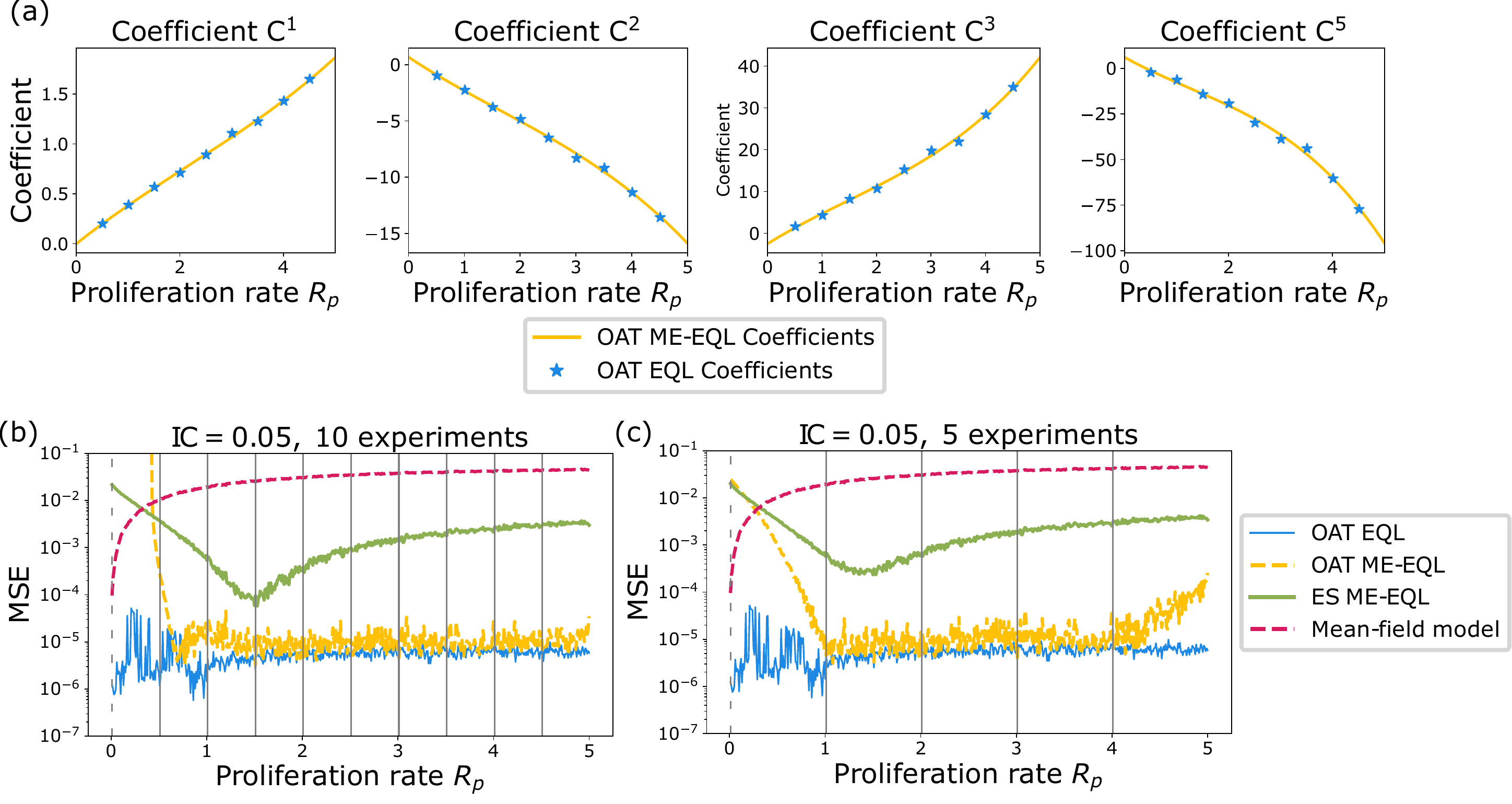}
\caption{Comparison of the generalizability of learned equations using mean-field model (Eq~\eqref{eqn:MFM}), OAT ME-EQL, and ES ME-EQL for the ABM Model with IC = 0.05. Panel (a) displays the interpolated coefficients for the OAT ME-EQL method (yellow) using 10 OAT EQL learned parameters (blue stars). Panels (b), (c) display the MSE between data and recovered models for learning from a maximum of 10 $R_p$ values (b) and 5 $R_p$ values (c). The results from the non-generalized OAT EQL for each separate $R_p$ value are shown in blue, the OAT ME-EQL model with interpolated coefficients is shown in yellow dashes, and the ES ME-EQL learning is shown in green solid lines. The mean-field approximation (Eq~\eqref{eqn:MFM}) is depicted in red dashes. Gray vertical bars indicate the small set of $R_p$ values from
which OAT ME-EQL coefficients were learned from. Dashes indicate that OAT ME-EQL
did not include the dataset corresponding to that $R_p$ value, since this framework did not learn the most popular model at that parameter value. For very small $R_p$ values, the mean-field approximation results in the lowest MSE for all the generalizable models. However, for all other $R_p$ values, the OAT ME-EQL approach outperforms the mean-field model and ES ME-EQL approaches in generalizability.} 
\label{fig:ABM_interp_0.05} 
\end{figure}

We now test the ability of the learned models to predict ABM population data at parameter values not used in learning. Fig~\ref{fig:ABM_interp_0.05} displays the results for all models when learning from 10 experiments (i.e., 10 $R_p$ values) and 5 experiments for IC = 0.05. When using 10 $R_p$ values for interpolation, we observe that the interpolated coefficients for OAT ME-EQL are nonlinear (Fig~\ref{fig:ABM_interp_0.05}a). Of the three ME methods, the OAT ME-EQL learns models with the lowest MSE values for most $R_p$ values, although the mean-field DE model performs best for small $R_p$ values (Fig~\ref{fig:ABM_interp_0.05}b). Similar results are obtained when using 5 experiments, but we find that the OAT ME-EQL method learns DE models with slightly worse MSE for $R_p<1$ and $R_p>4$ (Fig~\ref{fig:ABM_interp_0.05}c). This indicates the method may be sensitive to the exact $R_p$ values for which data is available.

\begin{figure}[t!]
\centering
\includegraphics[width=
0.9\textwidth]{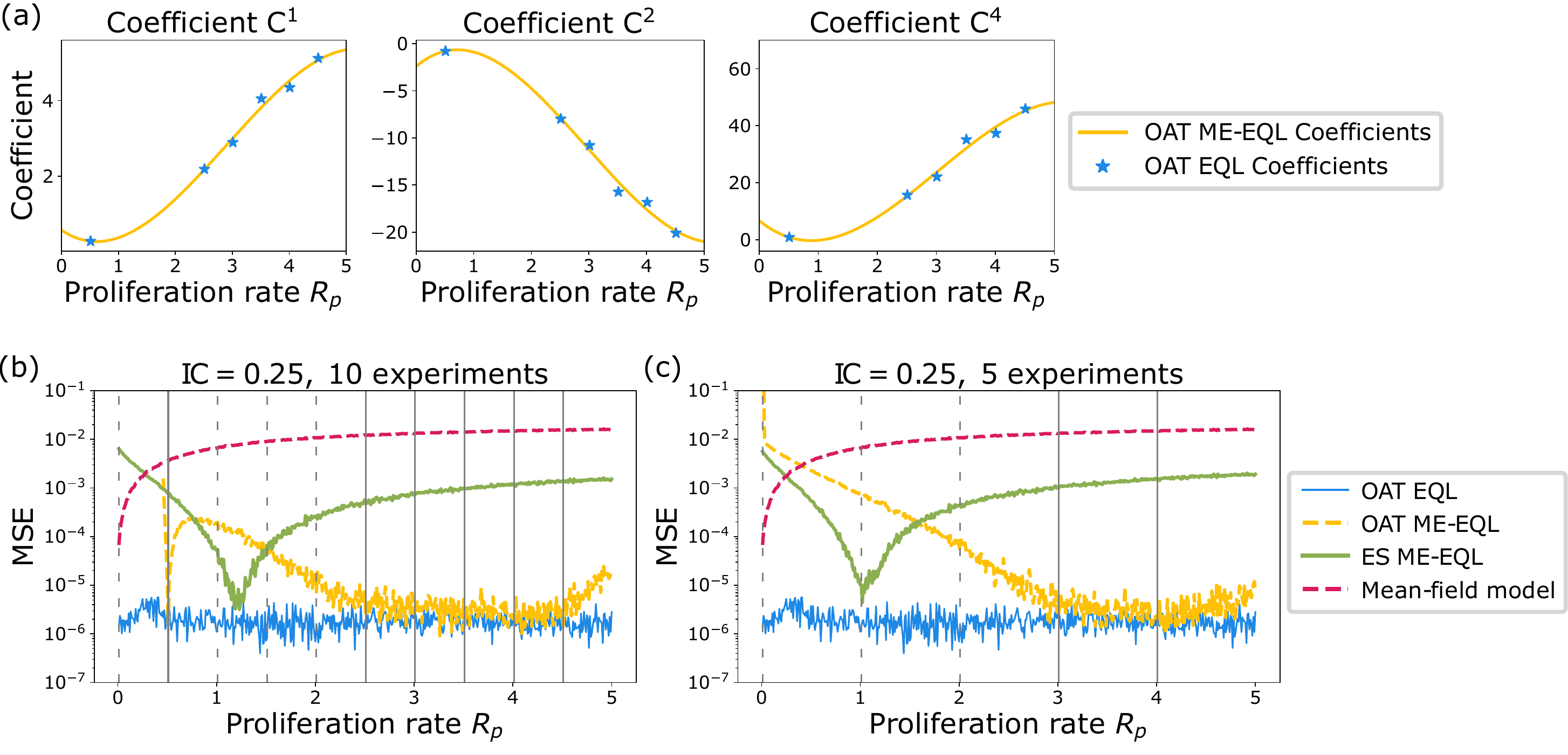}
\caption{Comparison of the generalizability of learned equations using the mean-field model approximation (Eq~\eqref{eqn:MFM}), OAT ME-EQL, and ES ME-EQL for the ABM Model with IC = 0.25. Panel (a) displays the interpolated coefficients for the OAT ME-EQL method (yellow) using a maximum of 10 OAT EQL learned parameters (blue stars). Panels (b), (c) display the MSE between data and recovered models for learning from a maximum of 10 $R_p$ values (b) and 5 $R_p$ values (c). The results from the non-generalized OAT EQL for each separate $R_p$ value are shown in blue, the OAT ME-EQL model with interpolated coefficients is shown in yellow dashes, and the ES ME-EQL learning is shown
in green solid lines. The mean-field approximation (Eq~\eqref{eqn:MFM}) is depicted in red dashes. Gray vertical bars indicate the small set of $R_p$ values from which OAT ME-EQL coefficients were learned from. Dashes indicate that OAT ME-EQL did not include the dataset corresponding to that $R_p$ value, since this framework did not learn the most popular model at that parameter value. When examining generalizability, at very small $R_p$ values, the mean-field approximation results in the lowest MSE. However, for all other $R_p$ values, OAT ME-EQL outperforms the mean-field model and ES ME-EQL in generalizability. In contrast to the IC=0.05 case (Fig~\ref{fig:ABM_interp_0.05}), there is more variation in learned models for OAT EQL, and thus, the OAT ME-EQL method rejects more learned models (using only 6 out of a maximum of 10) for interpolation.}
\label{fig:ABM_interp_0.205}
\end{figure}

Fig~\ref{fig:ABM_interp_0.205} illustrates the ability of the learned models to generalize to out-of-sample parameters for the ABM data with IC = 0.25. In general, there is less consistency in the learned models over the 10 equally-spaced $R_p$ values considered: only 6 of the 10 experiments share the same model structure in the OAT EQL approach. In particular, the most popular model is consistently learned for larger $R_p$ values ($R_p > 2.5$). As a result, the OAT ME-EQL approach learns predictive models for larger $R_p$ values, but generates worse predictions for smaller $R_p$ values. When learning from 5 experiments, the OAT EQL approach only recovers the most popular model for 2 $R_p$ values. The OAT ME-EQL approach learns a model whose prediction deteriorates for $R_p$ values outside the range of these two values. Still, the OAT ME-EQL approach outperforms the ES ME-EQL approach in generalizability, with the exception of select small $R_p$ parameter ranges. 

\subsection*{Can we infer $R_p$ from ABM data?}
\label{sec:inferRP}
Limited sampling in experimental biology, particularly in spatiotemporal cellular and ecological systems, poses significant challenges for using mathematical models to investigate mechanisms generating the observed dynamics. This challenge arises from the need to estimate model parameters from sparse and noisy data \cite{harrison2018impact, lagergren_biologically-informed_2020}. Therefore, it is essential to assess whether generalizable surrogate models of ABMs can be learned from a limited number of simulations or experiments. Here, we investigate whether the multi-experiment learning frameworks proposed can yield ODE models that retain predictive accuracy when applied to out-of-sample experimental conditions. 

We investigate the accuracy of each of the three parameterized models in estimating the parameter $R_p$ that generates a single noisy ABM simulation. To assess how this accuracy varies with $R_p$, we simulated the ABM for 50 different experiments for $R_p$ values between 0.01 and 5.0 for both ICs of 0.05 and 0.25. We estimate the value of $R_p$ that generated a single noisy ABM dataset $C_d(t)$ using a DE model $C(t;R_p)$ by minimizing:
\begin{equation}
    \hat{R}_p = \arg\min_{R_p\in\mathbb{R}} \sum_{i=1}^N (C_d(t_i) -C(t;R_p) )^2.
\end{equation}
Once we have obtained an estimate $\hat{R}_p$, we estimate its relative error as:
\begin{equation}
    \text{Relative $R_p$ Error} = \left| \dfrac{\hat{R}_p - R_p}{R_p} \right|.
\end{equation}
To understand the uncertainty of our $R_p$ estimate, we calculate 10 separate ABM datasets at each $R_p$ value; we then estimate $R_p$ for each of the 10 noisy datasets. This results in 10 $R_p$ estimates for each $R_p$ value and initial condition, and we report the mean and standard deviation of errors of these values.

\begin{figure}[t!]
    \centering
    \includegraphics[width=0.95\linewidth]{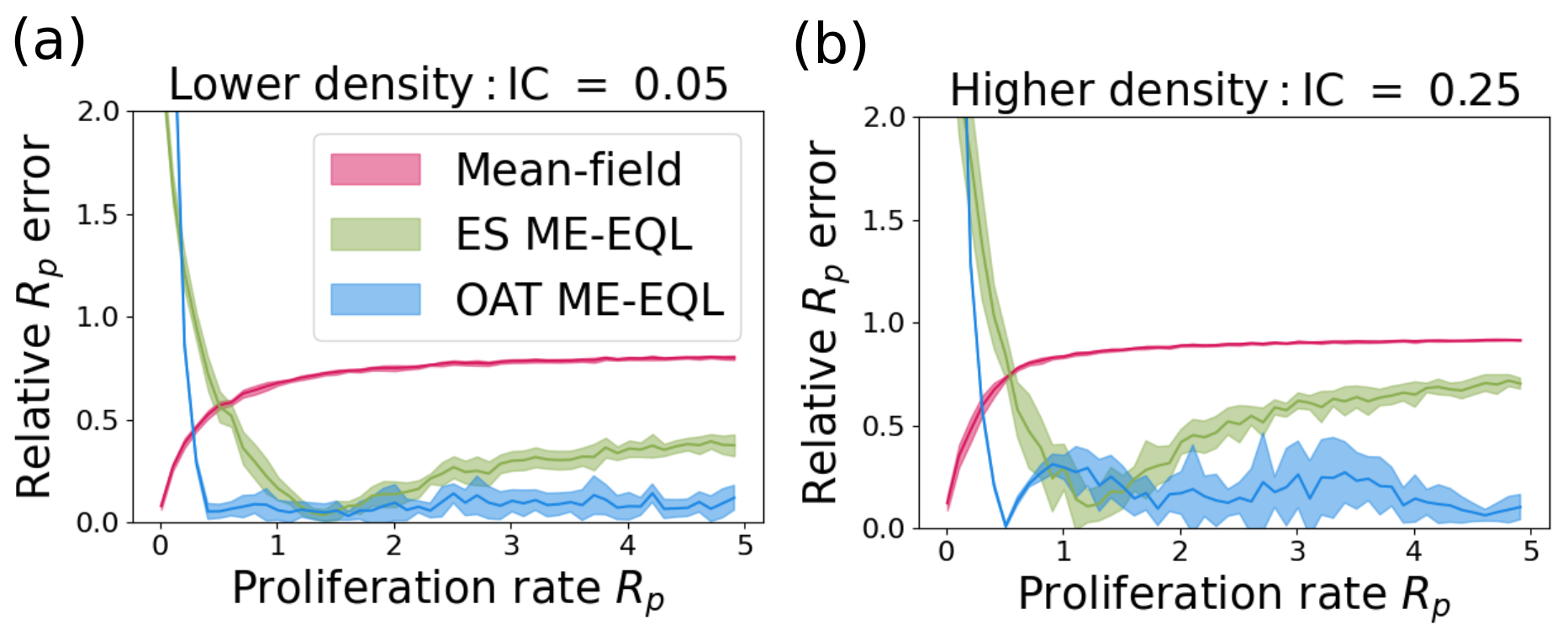}
    \caption{Error in learning ABM parameter $R_p$ from a single out-of-sample ABM simulation using the mean-field DE, ES ME-EQL, and OAT ME-EQL models for (a) IC = 0.05 and (b) IC= 0.25. Mean-field results are shown in magenta, ES ME-EQL is shown in green, and OAT ME-EQL is displayed in blue. Solid bars represent the mean error over 10 ABM simulations, and the shaded area represents one standard deviation about the mean. For small values of $R_p$, the mean-field model provides the best approximation of $R_p$. For IC = 0.05, OAT ME-EQL approximations generally perform the best for larger $R_p$ values. For IC = 0.25, there are regions in $R_p$ parameter space for which OAT ME-EQL and ES ME-EQL produce similar errors, but for larger $R_p$ values, OAT ME-EQL produces better fits.  }
    \label{fig:r3result}
\end{figure}

We compare the performance of the mean-field DE model and the learned equations from the OAT ME-EQL and ES ME-EQL pipelines from \nameref{sec:results_agentbased} that we learned from 10 $R_p$ values (Fig~\ref{fig:r3result} and Table \ref{tab:R3_tab}). For both initial conditions, the mean-field DE results in the lowest relative error for very small values of $R_p$, and the two ME-EQL learned model poorly estimate these $R_p$ values. For $R_p$ values above 0.33, however, the mean-field DE obtains higher error values than the two ME-EQL approaches. The OAT ME-EQL learned model achieves the most accurate estimates for most values of $R_p$ above 0.5, although the ES ME-EQL learned model achieves comparable estimates for values of $R_p$ between 1 and 2.

\begin{table}[]
    \centering
    \begin{tabular}{|c|c|c|c|}
    \hline
          & $R_p=0.01$ & $R_p = 2.51$ & $R_p=4.91$  \\
          \hline
         Mean-field & \textbf{0.081} & 0.779 & 0.802 \\\hline
         ES ME-EQL  &  2.273 & 0.270 & 0.377  \\ \hline
         OAT ME-EQL & 53.464 & \textbf{0.142} &  \textbf{0.121}\\
         \hline
    \end{tabular}
    \caption{Relative $R_p$ error values for various $R_p$ values for all three parameterized models for IC=0.05. Bold values denote the lowest relative errors (and, in turn, best parameter prediction) for each presented $R_p$ value.}
    \label{tab:R3_tab}
\end{table}

\section*{Discussion and conclusions}\label{sec:discussion}

Agent-based models are a natural means of describing spatiotemporal dynamics in many biological systems, yet the stochastic and parameter-heavy structure of these models presents challenges for inference and analysis. This motivates the development of non-spatial, population-level models to approximate ABMs, whether derived analytically from first principles or through equation learning. In the case of the birth--death--migration dynamics that we considered here, a mean-field model (under the case of well-mixing) is well known, but this ODE does not agree well with ABM simulations in parameter regimes where spatial correlations are significant. At the other extreme, the traditional EQL approach of learning a non-parameterized model for each experiment faces issues related to generalizability and out-of-sample prediction. Motivated by these challenges, we propose two data-driven methods for identifying generalizable and parameterized population-level models from multiple biological experiments. \edit{Our intent is to compare and contrast the two methods for this ABM system and understand the distinct challenges associated with each.} Our work represents an intermediate framework between deriving a mean-field model from ABM rules and learning a single model for each ABM parameter set.

Specifically, to help enhance generalizability in EQL, we considered two ways of learning ODE models from simulations of a birth--death--migration ABM or its associated mean-field model under a range of proliferation rates. Our two methods---OAT and ES ME-EQL---differ in how they incorporate information from multiple experiments. In the OAT ME-EQL approach, we performed equation learning $m$ times for $m$ proliferation rate values, resulting in $m$ models. Post-learning, we interpolated the coefficients for each common learned term as a function of the ABM proliferation rate, resulting in a single model. In our ES approach, we instead embedded the proliferation rate directly in our library and used all $m$ ABM datasets jointly for learning a single equation. Both methods led to parameterized models that closely fit our data. Moreover, in the case of ABM data with strong spatial correlations, our learned models generally outperformed the mean-field model. We evaluated our methods through out-of-sample prediction and found that five experiments (i.e., datasets corresponding to five proliferation rates) were sufficient to learn generalizable models that provide accurate predictions across unobserved proliferation rates.

Population size data are often more readily available than spatial data in complex biological systems, and this makes inferring parameters in a population-level model (as a surrogate for its corresponding ABM) attractive. However, mean-field models are not known for many biological systems, or they may rely on such strong simplifications that insight into parameters at the mean-field level does not translate into insight at the ABM level. OAT and ES ME-EQL help address these challenges and provide a means of estimating agent-based parameters from population data. We found that our methods could be used to recover ABM parameter values by fitting our learned parameterized models to population density in time from noisy ABM simulations. This is a major benefit of learning generalizable models that establish a map between ABM and ODE parameters. Moreover, while a mean-field model is known for the dynamics that we considered, we recovered more accurate ABM proliferation rates using our learned models than using the mean-field model, except in settings with weak spatial structure. This opens up many exciting directions. In the future, it will be interesting to apply our approach to inference in more complicated ABMs for which mean-field models are not known, as well as to estimate rates directly from biological data, similar to the SMoRe ParS framework \cite{jain2022smore,bergman2024connecting}. For example, Gerlee et al. \cite{gerlee2022autocrine} used a spatial ABM of tumor initiation to study how autocrine signaling can generate Allee effects in 2D \emph{in vitro} glioblastoma cell line time series data, while Malik et al. \cite{malik2025mathematical} fit reaction–advection–diffusion PDE models to growth curves of multicellular tumor spheroids from patient-derived glioblastoma cell lines to quantify inter-patient and intra-tumor heterogeneity. Our multi-experiment equation learning approach could be applied to such experimental time series to learn differential equations across multiple patient cell lines simultaneously, thereby enhancing the generalizability of the inferred equations governing proliferation, migration, and signaling feedback in glioblastoma cell populations. One could also compare our methods for ABM parameter recovery to inference based on more complicated ODE models that account for spatial correlations in time \cite{baker_correcting_2010,nardini_learning_2020}.

Overall, our two ME-EQL methods face different challenges. While OAT ME-EQL often resulted in lower mean square errors than did ES ME-EQL, its use of interpolation can be a limitation. In particular, because we interpolate after restricting to the most commonly-learned model structure, OAT ME-EQL may be unreliable when there is high variability in the terms learned. 
On the other hand, one limitation of ES ME-EQL is its dependence on knowledge of an appropriate library. We included terms with linear dependence on the proliferation rate $R_p$ to match the complexity in our two methods and 
because birth--death--migration dynamics give rise to a mean-field model with terms that depend linearly on $R_p$. If there is no information on how model terms depend on the parameters of interest, the library in ES ME-EQL would grow. In this sense, OAT ME-EQL 
requires less knowledge of the underlying biological dynamics. In the future, it will be interesting to learn from ABM dynamics when a mean-field model exists but our libraries do not include the mean-field model terms. We expect that OAT ME-EQL may be less prone to bias from a poorly chosen library in this setting. Moreover, by enforcing a unified model structure, ES ME-EQL assumes that a single model under varying parameters can explain our biological system. In settings where it is not known if this is the case, 
OAT ME-EQL could be used first to help identify whether or not a single model applies. After distinguishing regimes of system behavior, we could then apply either OAT ME-EQL or ES ME-EQL to learn a model in each regime.

Both of our ME-EQL frameworks may benefit from further investigation into their implementation and optimization. We used LASSO for sparse regression and AIC for model selection due to their simplicity and wide usage in the EQL literature \cite{tibshirani_regression_1996, mangan2016inferring}. Other options for sparse regression include ridge regression \cite{marquardt_ridge_1975}, greedy methods \cite{zhang_adaptive_2011}, or sparse relaxed regularized regression (SR3) \cite{zheng_unified_2019}; see \cite{bertsimas_sparse_2020} for a comprehensive review. While LASSO is intended to perform sparse regression, it may lead to parameter shrinkage without thresholding small parameters to zero \cite{van_de_geer_adaptive_2011}. To combat this, methods such as sequentially thresholded least squares (STLSQ) perform multiple iterations of ridge regression and parameter thresholding \cite{rudy_data-driven_2017, kang_ident_2019, schneider_ensemble_2022}, either based on parameter values themselves or the magnitude of the associated terms \cite{Messenger2021weakGalerkin, kang_ident_2019}. In contrast to LASSO, which requires the selection of a single hyperparameter for regularization strength, methods like STSQL require additional hyperparameters to implement thresholding. In future work, methods such as Bayesian optimization \cite{snoek_practical_2012} could be explored to overcome the computational bottleneck associated with selecting hyperparameters in higher dimensions. \edit{In our ME-EQL frameworks, we also used domain knowledge to avoid large coefficient values in the learned models that would be unrealistic for the system studied here. In future work, it would be interesting to determine how the results change with no prior knowledge, and whether this impacts the comparison of the learned models from OAT and ES ME-EQL. }


There are many ways to build on our study, and we highlight several here. For example, it will be interesting to further consider the role of noise in EQL. By first learning from mean-field model data with varying noise levels, we determined how our two methods perform when there is a known (e.g., \textit{correct}) model to learn. Surprisingly, we found that, even in the case of noise-free data, single-experiment EQL and our two ME-EQL methods did not always learn the terms in the correct underlying model. In the future, it would be interesting to investigate the robustness of ME-EQL to the number of ABM simulations, as increasing the number of simulations would result in smoother data but be more computationally expensive. Learning from fewer ABM runs would also be of interest, given the typical noise levels in available experimental data. Our successful inference of parameter $R_p$ from a single, noisy, and out-of-sample ABM simulation in Fig~\ref{fig:r3result} suggests that learning from fewer datasets could also yield ODE models that are generalizable and predictive. We could also consider improved methods for numerically calculating derivatives from noisy data.  While we chose to base our work on SINDy \cite{brunton2016discovering}, related methods such as weak SINDy \cite{messenger2021weak,Messenger2021weakGalerkin} could be used to overcome challenges related to approximating the time derivatives needed for equation learning. Incorporating weak SINDy into our ES ME-EQL approach would be a potential avenue for future work to improve upon the results presented here.

Another natural extension of our approach would be to consider multiple independently varying parameters. In this setting, OAT ME-EQL could scale relatively well, since models could still be learned separately for each parameter combination, and coefficients interpolated to predict dynamics for unobserved parameter sets. ES ME-EQL could also be extended, but including multiple parameters in the candidate library would increase its size combinatorially, raising computational costs and reducing interpretability.


Finally, another exciting direction for future work is refining our choice of ABM parameter values to determine the most informative experiments for ME-EQL. This would inform experimental design, but is also related to sensitivity analysis, which could be used to show that our learned model is more sensitive at low proliferation rates and suggest that more experiments should be performed there. More broadly, the major benefit of learning generalizable, parameterized models is that they are amenable to traditional, powerful approaches such as bifurcation analysis, optimal control, and uncertainty quantification. We expect that combining equation learning with such classic modeling approaches will shed new light on biological systems in the future.

\section*{Supporting information}

\paragraph*{S1 Text.}
\label{S1_file}
{\bf Calculating the optimal regularization hyperparameter $\lambda$.} Details on methods to calculate the optimal regularization hyperparameter $\lambda$ and  the procedure used to select the final model.

\paragraph*{S2 Text.}
\label{S2_file}
{\bf Algorithmic differences between OAT ME-EQL and ES ME-EQL.} Algorithms for the OAT ME-EQL and ES ME-EQL approaches.

\paragraph*{S1 Fig.}
\label{S1_Figure}
Example plots of the hyperparameter $\bar{\lambda}$ plotted against the AIC scores of $R_p=1$ (left) and $R_p=5$ (right). Optimal $\lambda$ selected in red square. Text indicates the learned model structure  at each jump in the plot.

\paragraph*{S2 Fig.}
\label{S2_Figure}
{\bf{Example model simulations highlighting how initial conditions affect information content.}} Snapshots of ABM simulations of birth, death, and migration dynamics at different timepoints for initial conditions with $5$\% and $25$\% of sites occupied, respectively, and corresponding mean population sizes in time across $25$ ABM simulations.

\paragraph*{S1 Table.}
\label{S1_Table}
{\bf Models learned using ME-EQL for mean-field model data.} Models learned using ME-EQL methods for the mean-field model data with initial conditions 0.05 and 0.25, and for noise levels $\sigma = 0\%$ and $\sigma = 0.25\%$.

\paragraph*{S2 Table.}
\label{S2_Table}
{\bf Models learned using ME-EQL for ABM data.} Models learned using ME-EQL methods for the ABM data with initial conditions 0.05 and 0.25.

\section*{Data and code availability} All code and data for this study is publicly available at \href{https://github.com/johnnardini/ME-EQL}{https://github.com/johnnardini/ME-EQL}.

\section*{Acknowledgments}
This project began during a SQuaRE at the American Institute for Mathematics. The authors thank AIM for providing a supportive and mathematically rich environment. The authors also acknowledge use of the ELSA high performance computing cluster at The College of New Jersey for conducting the research reported in this paper. This cluster is funded in part by the National Science Foundation under grant numbers OAC-1826915 and OAC-2320244. This work was funded in part by the National Science Foundation under grant numbers DMS-2327836, DMS-2342344, and DMS-2424748 to KBF; the National Institute of Allergy and Infectious Diseases (NIAID) of the National Institutes of Health (1U54AI191253-01, to K.B.F. and M.V.C.).


\begin{appendices}
\section{Supplementary Methods}

\subsection*{S1 Text: Calculating the optimal regularization hyperparameter $\lambda$}
\label{s1text}
When optimizing $\hat{\xi}$ in Eq. 7, the hyperparameter $\lambda$ governs the tradeoff between model complexity and fit. Since we do not know \textit{a priori} how complex the model should be, we perform cross-validation to determine the optimal regularization parameter, $\lambda$, for the LASSO algorithm, using the pySINDy package for SINDy implementation \cite{desilva2020,Kaptanoglu2022}. 

We perform a grid search over $\lambda_j = 10^{-1},...,10^{-9}$ with 100 equi-log-spaced values. For each value $\lambda_j$, we randomly select 10 train-test splits (each with the standard 80\% training and 20\% testing data) of $\frac{dC_d(t)}{dt}$. For each split, we solve Eq. 7 using LASSO with the specified library to obtain the optimal coefficients $\hat{\xi_k}, k=1,...,10$. For each optimal coefficient $\hat{\xi_k}$, we then forward solve Eq. 6 and calculate the sum of squared errors (SSE) between the forward solution of the learned differential equation model and the test data. \edit{Using this information, we calculate the Akaike Information Criteria (AIC) score for each train-test split. We note that using AIC or variants such as the corrected AICc for model selection is an established practice in the literature \cite{mangan2017model}. Since our dataset is not sparse in time, the correction is not necessary and we use the standard AIC score.}

Each value $\lambda_j$ is quantified with a final AIC score which is the average AIC over all 10 train-test splits. Once all AIC values are calculated for all $\lambda$, we select the $\lambda_j$ for which the mean AIC is minimal. However, we found that when $\lambda$ was too small, selected models had many nonzero terms with large non-interpretable coefficients. The smaller the value of $\lambda$, the larger the oscillating coefficients. 
Therefore, we defined the \textit{lower bound} for the optimal $\lambda$ as the $\lambda$ with the minimum AIC score. 

Because we do not expect large coefficients due to our domain knowledge, we require that the optimal $\lambda$ selected must have every recovered model in the train-test split have all model coefficients below a threshold value, which we set to 100 for OAT ME-EQL and 20 for ES ME-EQL (these two values correspond to each other because in the ES ME-EQL approach, each coefficient is multiplied by $R_p$, which varies between 0 and 5). If we find that the value of $\lambda$ with the minimum AIC score contained at least one model for which any coefficient is above the specified threshold, we increased the value of $\lambda$ until we find one that met this criterion. \\

\begin{figure}[h]
    \centering
    \includegraphics[width=0.45\linewidth]{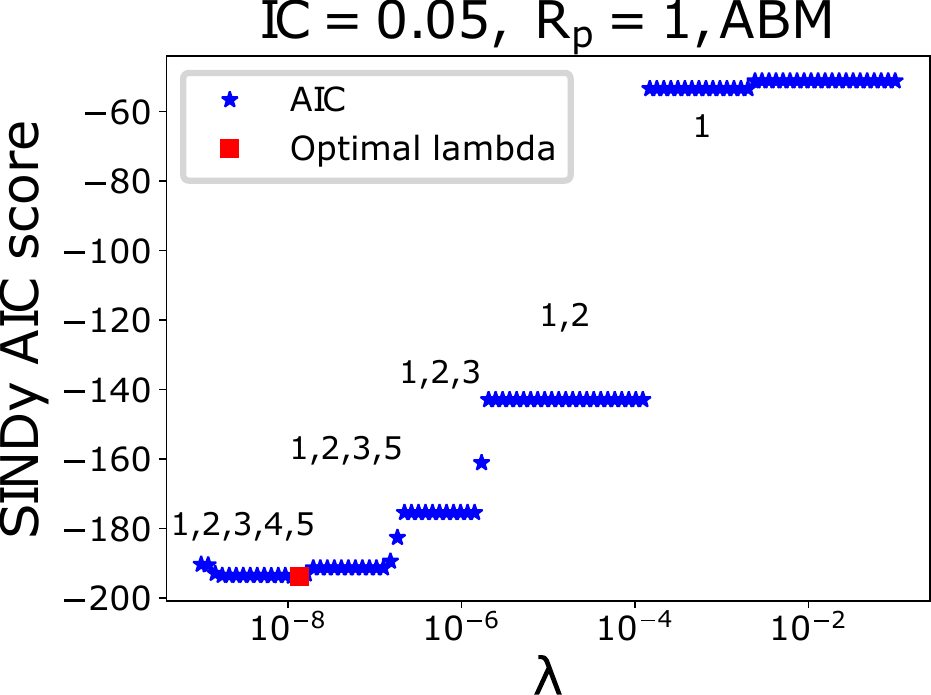}
  \includegraphics[width=0.45\linewidth]{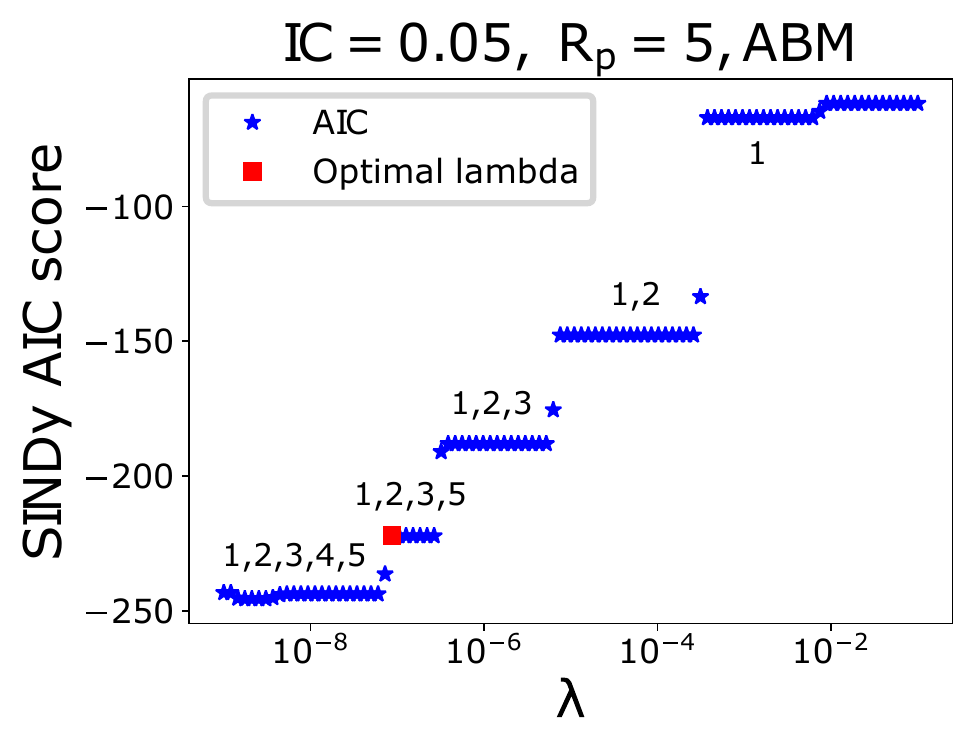}
    \caption{\edit{Example plots of the hyperparameter $\bar{\lambda}$ plotted against the AIC scores of $R_p=1$ (left) and $R_p=5$ (right). Optimal $\lambda$ selected in red square. Text indicates the learned model structure  at each jump in the plot.}}
    \label{fig:lambdavsAIC}
\end{figure}

\edit{We display an example plot of $\bar{\lambda}$ and AIC values in \Cref{fig:lambdavsAIC} for $R_p=1$ (left) and $R_p=5$ (right). For each case, the plots appear step-like, indicating different model structures are being selected at each jump in the plot (labeled with which terms are in the final selected model structure). Note that the final selected $\lambda$ is not always the one that generates the lowest mean AIC score, but instead must also satisfy the second criteria in which none of the learned coefficients can be greater in magnitude than 100 for any of the 10 test-train splits. }

\subsection*{Selecting the final model}
Once the optimal $\lambda$ is selected, we examine the 10 test-train splits to determine the most popular learned model structure. Of the 10 models that contain the \edit{most popular learned} model structure, we average the coefficient values to find $\bar{\xi}$. \edit{To find the final model coefficients} \editmath{$\xi_{f}$}, \edit{we re-optimize the parameters to the most popular learned model structure using all of the data. We use a Nelder-Mead algorithm and} \editmath{$\bar{\xi}$} \edit{is  used as the initial parameter guess.} \\


\newpage

\subsection*{S2 Text: Algorithmic differences between OAT ME-EQL and ES ME-EQL}

The one-at-a-time (OAT) and embedded structure (ES) variants of ME-EQL differ in how they handle data across parameter values during both hyperparameter tuning and model selection. A detailed description of the methodology for hyperparameter tuning and model selection can be found in the Supplementary Material, and the corresponding algorithms for each of the two equation learning approaches are summarized in Algorithms 1 and 2 below. In the OAT approach, each dataset corresponding to a specific parameter value is treated independently: for each $R_p$, a separate grid search over $\lambda$ is performed, leading to potentially different optimal regularization parameters and selected models across parameter settings. After model selection via AIC-based majority vote, each model is re-optimized individually and the final OAT ME-EQL model is constructed by interpolating coefficients across the most commonly-retained models. In contrast, the ES ME-EQL approach treats the entire set of datasets jointly by embedding parameter values directly into the function library. As a result, a single global $\lambda$ is selected through cross-validation over all parameterized datasets, and the model structure is determined via majority vote across the aggregate train-test splits. This unified treatment enforces structural consistency across the parameter space and produces a single, global model fit to all the data. Thus, OAT ME-EQL is better suited for uncovering local structures and heterogeneity, while ES ME-EQL emphasizes global coherence and structural generalizability.

\begin{algorithm}[H]
\caption{One-at-a-Time (OAT) ME-EQL Model}
\begin{algorithmic}[1] 

    \State \textbf{Input:}  
    \Statex \hspace{1em} Set of parameters $P = \{R_p\}$  
    \Statex \hspace{1em} Data for each parameter: $D = \left\{ C_{d,R_p} \right\}$  
    \Statex \hspace{1em} Regularization set: $\Lambda = \{\lambda_j\}$
    
    \State \textbf{Final Output:} OAT Model

    \ForAll{$R_p \in P$} 
        \ForAll{$\lambda_j \in \Lambda$} 
            \State Determine $\hat{\xi}_{R_p, \lambda_j}$ by minimizing Equation (7) in the main text
            \State Record AIC score 
        \EndFor
        \vspace{4pt}
        \State Select $\lambda^*$ with minimum AIC score and no learned coefficient magnitudes exceeding 100.
        \State Select model structure by majority vote across test-train splits. 
        \State Re-optimize model parameters using data for $R_p$.
    \EndFor

    \vspace{4pt}
    \State Retain $R_p$ values and optimized models with the majority learned structure.  
    \State Determine final OAT-model by interpolating coefficients for each term in the majority model over $R_p$ values.   

\end{algorithmic}
\end{algorithm}

\vspace{10pt} 

\begin{algorithm}[H]
\caption{Embedded Structure (ES) ME-EQL Model}
\begin{algorithmic}[1]

    \State \textbf{Input:}  
    \Statex \hspace{1em} Set of parameters $P = \{R_p\}$  
    \Statex \hspace{1em} Data for all parameters: $D = \left\{ [C_{d,R_{p_1}},C_{d,R_{p_2}},\dots, C_{d,R_{p_N}}] \right\}$  
    \Statex \hspace{1em} Regularization set: $\Lambda = \{\lambda_j\}$

    \State \textbf{Final Output:} ES Model

    \ForAll{$\lambda_j \in \Lambda$} 
        \State Determine $\hat{\xi}_{\lambda_j}$ by minimizing  minimizing Equation (7) in the main text
        \State Record AIC score
    \EndFor

    \vspace{4pt}
    \State Select  $\lambda^*$ with minimum AIC score and no learned coefficient magnitudes exceeding 20.
    \State Select model by majority vote across test-train splits.
    \State Determine final ES-model: Re-optimize using $D$. 

\end{algorithmic}
\end{algorithm}

\newpage

\newpage 

\begin{figure}[h!]
\includegraphics[width=\textwidth]{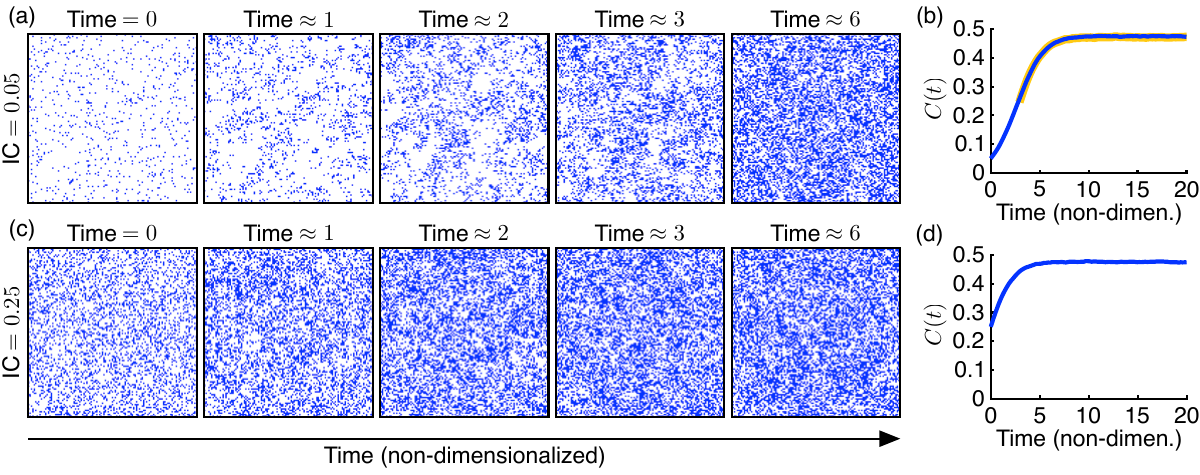}
\caption{\edit{Example model simulations highlighting how initial conditions affect information content. Panels (a) and (c) each present snapshots of an ABM simulation of birth, death, and migration dynamics at different timepoints for initial conditions with $5$\% and $25$\% of sites occupied, respectively. (Blue denotes occupied sites, while white denotes unoccupied sites.) Similarly, panels (b) and (d) show the corresponding mean population sizes in non-dimensionalized time for IC $= 0.05$ and IC $=0.25$, respectively, averaged across $25$ ABM simulations each. Here we set $R_p = 0.1$, $R_d = 0.05$, and $R_m = 1$, and consider a $120 \times 120$ square lattice. The yellow curve in panel (b) is a copy of the mean population-size curve from panel (d), shifted in time and shortened to roughly align with the latter portion of the $C(t)$ curve in panel (b). This yellow curve serves to highlight that changing the ABM initial conditions allows us to control the amount of information content in our data. In particular, using IC $=0.25$ effectively means that we observe less of the ABM time dynamics than we do when IC $=0.05$.}}
\end{figure}


\begin{sidewaystable}[h!]
\centering 
\caption{Models learned using ME-EQL methods for the mean-field model data with initial conditions 0.05 and 0.25, and for noise levels $\sigma = 0\%$ and $\sigma = 0.25\%$. Coefficients are displayed and rounded to at most 3 decimal places. The bolded equation is the ground-truth, correct mean-field model. }
\begin{tabular}{|l|l|l|l|}
\hline
\textbf{Setting}                                                    & \textbf{\begin{tabular}[c]{@{}l@{}}Experiments\\ used\end{tabular}} & \textbf{OAT ME-EQL}                               & \textbf{ES ME-EQL}                          \\ \hline
\multirow{3}{*}{IC = 0.05, $\sigma = 0\%$}                          & 500                                                                 & $\boldsymbol{dC/dt = 0.50 R_p C-1.00R_pC^2}$          & $dC/dt = 0.50 R_p C-1.00R_pC^2$             \\ \cline{2-4} 
                                                                    & 10                                                                  & $dC/dt = 0.50 R_p C-1.00R_pC^2$                   & $dC/dt = 0.50 R_p C-1.00R_pC^2$             \\ \cline{2-4} 
                                                                    & 5                                                                   & $dC/dt = 0.50 R_p C-1.00R_pC^2$                   & $dC/dt = 0.50 R_p C-1.00R_pC^2$             \\ \hline
\multicolumn{1}{|c|}{\multirow{3}{*}{IC = 0.05, $\sigma = 0.25\%$}} & 500                                                                 & $dC/dt = 0.50 R_p C-(1.00R_p+0.001)C^2$           & $dC/dt = 0.50 R_p C-1.02R_pC^2+0.02R_p C^3$ \\ \cline{2-4} 
\multicolumn{1}{|c|}{}                                              & 10                                                                  & $dC/dt = 0.499 R_p C-(0.998R_p-0.001)C^2$         & $dC/dt = 0.50 R_p C-1.00R_pC^2$             \\ \cline{2-4} 
\multicolumn{1}{|c|}{}                                              & 5                                                                   & $dC/dt = (0.5 R_p-0.002) C-(1.001R_p-0.005)C^2$   & $dC/dt = 0.50 R_p C-1.00R_pC^2$             \\ \hline
\multirow{3}{*}{IC = 0.25, $\sigma = 0\%$}                          & 500                                                                 & $dC/dt = 0.50 R_p C-1.00R_pC^2$                   & $dC/dt = 0.49 R_p C-0.97R_pC^2-0.03R_p C^3$ \\ \cline{2-4} 
                                                                    & 10                                                                  & $dC/dt = 0.50 R_p C-1.00R_pC^2$                   & $dC/dt = 0.5 R_p C-0.98R_pC^2-0.02R_p C^3$  \\ \cline{2-4} 
                                                                    & 5                                                                   & $dC/dt = 0.50 R_p C-1.00R_pC^2$                   & $dC/dt = 0.5 R_p C-1.02R_pC^2+0.03R_p C^3$  \\ \hline
\multirow{3}{*}{IC = 0.25, $\sigma = 0.25\%$}                       & 500                                                                 & $dC/dt = 0.501 R_p C-(1.001R_p+0.001)C^2$         & $dC/dt = 0.50 R_p C-1.01R_pC^2+0.01R_pC^3$  \\ \cline{2-4} 
                                                                    & 10                                                                  & $dC/dt = (0.50 R_p+0.003) C-(1.001R_p+0.006)C^2$  & $dC/dt = 0.51 R_p C-1.04R_pC^2+0.05R_pC^3$  \\ \cline{2-4} 
                                                                    & 5                                                                   & $dC/dt = (0.508 R_p-0.002) C-(1.016R_p-0.004)C^2$ & $dC/dt = 0.51 R_p C-1.05R_pC^2+0.06R_pC^3$  \\ \hline
\end{tabular}
\end{sidewaystable}


\begin{sidewaystable}[ht!]
\centering 
\caption{Models learned using ME-EQL methods for the ABM data with initial conditions 0.05 and 0.25. Coefficients are displayed and rounded to at most 3 decimal places. }
\begin{tabular}{|l|l|l|l|}
\hline
\textbf{Setting}                                 & \textbf{\begin{tabular}[c]{@{}l@{}}Experiments\\ used\end{tabular}} & \textbf{OAT ME-EQL}                                                                                                                                                                                                                                                                     & \textbf{ES ME-EQL}                                                                                    \\ \hline
\multirow{3}{*}{IC = 0.05}                       & 500                                                                 & \begin{tabular}[c]{@{}l@{}}$dC/dt = (-0.003 R_p^3 +  0.023 R_p^2 +  0.29 R_p + 0.067) C^1$ \\ $+ (0.06 R_p^3 -0.603 R_p^2 -1.286 R_p -0.409) C^2$ \\ $+ (-0.215 R_p^3 + 2.263 R_p^2 + 1.182 R_p + 1.116) C^3 $\\ $+ (0.538 R_p^3 - 6.3 R_p^2 + 2.262 R_p - 3.016) C^5$\end{tabular}     & \begin{tabular}[c]{@{}l@{}}$dC/dt = 0.36*R_pC - 2.34R_pC^2$\\ $+ 5.6R_pC^3 - 5.0R_pC^4$\end{tabular}  \\ \cline{2-4} 
                                                 & 10                                                                  & \begin{tabular}[c]{@{}l@{}}$dC/dt = (0.006 R_p^3 - 0.038 R_p^2 +  0.415 R_p - 0.0005) C^1$\\ $+ (-0.075 R_p^3 + 0.372 R_p^2 – 3.286 R_p + 0.665) C^2$\\ $+ (0.261 R_p^3 – 1.142 R_p^2 + 8.09 R_p – 2.525) C^3$\\ $+ (-0.739 R_p^3 + 2.777 R_p^2 – 15.855 R_p + 6.165) C^5$\end{tabular} & \begin{tabular}[c]{@{}l@{}}$dC/dt = 0.4R_pC - 2.53R_pC^2$ \\ $+ 5.81R_pC^3 - 4.92R_pC^4$\end{tabular} \\ \cline{2-4} 
                                                 & 5                                                                   & \begin{tabular}[c]{@{}l@{}}$dC/dt = (-0.025 R_p^3 + 0.186 R_p^2 -  0.065 R_p + 0.29) C^1$\\ $+ (0.22 R_p^3 – 1.769 R_p^2 + 1.202 R_p – 1.896) C^2 $\\ $+ (-0.516 R_p^3 + 4.451 R_p^2 – 3.397 R_p + 3.707) C^3 $\\ $ + (0.716 R_p^3 – 7.55 R_p^2 + 4.693 R_p – 4.042) C^5$\end{tabular}  & \begin{tabular}[c]{@{}l@{}}$dC/dt = 0.42R_pC - 2.37R_pC^2$\\ $+ 3.92R_pC^3 - 3.84R_pC^5$\end{tabular} \\ \hline
\multicolumn{1}{|c|}{\multirow{3}{*}{IC = 0.25}} & 500                                                                 & \begin{tabular}[c]{@{}l@{}}$dC/dt = (0.004 R_p^3 + 0.074 R_p^2 +  0.79 R_p - 0.197) C^1$\\ $+ (-0.008 R_p^3 – 0.423 R_p^2 -2.652 R_p + 0.886) C^2 $\\ $+ (0.004 R_p^3 + 1.486 R_p^2 + 4.028 R_p – 2.12) C^4$\end{tabular}                                                               & $dC/dt = 0.34 R_p C-0.85R_pC^2$                                                                       \\ \cline{2-4} 
\multicolumn{1}{|c|}{}                           & 10                                                                  & \begin{tabular}[c]{@{}l@{}}$dC/dt = (-0.107 R_p^3 + 0.93 R_p^2 – 1.02 R_p + 0.568) C^1$\\ $+ (0.463 R_p^3 – 4.092 R_p^2 + 5.163 R_p – 2.421) C^2$\\ $+ (-1.235 R_p^3 + 11.218 R_p^2 – 16.917 R_p + 6.732) C^4$\end{tabular}                                                             & \begin{tabular}[c]{@{}l@{}}$dC/dt = 0.85 R_p C - 2.95R_pC^2$\\ $+ 5.24 R_p C^4$\end{tabular}          \\ \cline{2-4} 
\multicolumn{1}{|c|}{}                           & 5                                                                   & \begin{tabular}[c]{@{}l@{}}$dC/dt = ( 1.453 R_p – 1.48) C^1 $\\ $+ (-6. 041 R_p + 7.39) C^2 $\\ $+ (15.247 R_p - 23.776) C^4$\end{tabular}                                                                                                                                              & \begin{tabular}[c]{@{}l@{}}$dC/dt = 0.86 R_p C - 2.92R_pC^2$\\ $+ 4.94 R_p C^4$\end{tabular}          \\ \hline
\end{tabular}
\end{sidewaystable}





\clearpage

\newpage
\bibliographystyle{unsrt}
\bibliography{bibliography}

\end{appendices}

\end{document}